\documentclass{article}

\usepackage{microtype}
\usepackage{graphicx}
\usepackage{subfigure}
\usepackage{booktabs} 
\usepackage{amsmath}
\usepackage{amsthm}
\usepackage{pythonhighlight}
\usepackage{hyperref}
\usepackage{amssymb}
\usepackage{soul}
\usepackage{cleveref}
\usepackage{natbib}

\newcommand{\Exp}{\mathop{\mathbb{E}}}

\newcommand{\bs}[1]{\boldsymbol{#1}}

\usepackage[accepted]{arXiv}

\icmltitlerunning{An Interactive Framework for Finding the Optimal Trade-off in Differential Privacy}

\theoremstyle{plain}
\newtheorem{theorem}{Theorem}[section]

\theoremstyle{definition}
\newtheorem{definition}[theorem]{Definition}

\theoremstyle{remark}

\begin{document}

\twocolumn[
\icmltitle{\large
    \mbox{An Interactive Framework for Finding the Optimal Trade-off in Differential Privacy}
}



\icmlsetsymbol{equal}{*}

\begin{icmlauthorlist}
\icmlauthor{Yaohong Yang}{aalto}
\icmlauthor{Aki Rehn}{equal,HY}
\icmlauthor{Sammie Katt}{equal,aalto}
\icmlauthor{Antti Honkela}{HY}
\icmlauthor{Samuel Kaski}{aalto,manchester}

\end{icmlauthorlist}

\icmlaffiliation{aalto}{Department of Computer Science, Aalto University, Espoo, Finland}
\icmlaffiliation{HY}{Department of Computer Science, University of Helsinki, Helsinki, Finland}
\icmlaffiliation{manchester}{Department of Computer Science, University of Manchester, Manchester, United Kingdom}

\icmlcorrespondingauthor{Yaohong Yang}{yaohong.yang@aalto.fi}
\icmlcorrespondingauthor{Samuel Kaski}{samuel.kaski@aalto.fi}

\icmlkeywords{Machine Learning, ICML}

\vskip 0.3in
]

\printAffiliationsAndNotice{\icmlEqualContribution} 

\begin{abstract}

    Differential privacy (DP) is the standard for privacy-preserving analysis, and introduces a fundamental trade-off between privacy guarantees and model performance.
    Selecting the optimal balance is a critical challenge that can be framed as a multi-objective optimization (MOO) problem where one first discovers the set of optimal trade-offs (the Pareto front) and then learns a decision-maker's preference over them.
    While a rich body of work on interactive MOO exists, the standard approach -- modeling the objective functions with generic surrogates and learning preferences from simple pairwise feedback -- is inefficient for DP because it fails to leverage the problem's unique structure:
    a point on the Pareto front can be generated directly by maximizing accuracy for a fixed privacy level. 
    Motivated by this property, we first derive the shape of the trade-off theoretically, which allows us to model the Pareto front directly and efficiently.
    To address inefficiency in preference learning, we replace pairwise comparisons with a more informative interaction. 
    In particular, we present the user with hypothetical trade-off curves and ask them to pick their preferred trade-off.
    Our experiments on differentially private logistic regression and deep transfer learning across six real-world datasets show that our method converges to the optimal privacy-accuracy trade-off with significantly less computational cost and user interaction than baselines.

\end{abstract}

\section{Introduction} \label{sec:intro}
    Differential privacy (DP)~\cite{dwork2006calibrating} is a privacy-preserving paradigm that protects the sensitive data when training deep learning models.
    Its application introduces a fundamental trade-off between the strength of the privacy guarantee (e.g., privacy budget $\varepsilon$ in $(\varepsilon, \delta)$-DP~\cite{dwork2006calibrating}, $\rho$ in $\rho$-zCDP~\cite{bun2016concentrated, dwork2016concentrated} and $\mu$ in $\mu$-GDP~\cite{dong2022gaussian}), and the resulting model's performance (e.g., accuracy in classification tasks and validation loss in general deep learning models). 
    Critically, the performance of models is determined by the training hyperparameters.
    Therefore, navigating this trade-off presents a dual challenge for practitioners.
    First, the decision-maker must choose an appropriate value for the privacy level itself, a critical and context-dependent decision~\cite{dankar2012application,dwork2019differential}.
    Second, for a specific privacy level, one needs to perform hyperparameter optimization (HPO) over various training parameters to achieve the best possible accuracy. 
    While significant research has focused on the second challenge~\cite{pandanew,bu2025towards,koskela2023practical}, the first key question -- providing a systematic methodology for choosing the privacy level --has been largely overlooked. 
    Simply relying on conventional values (e.g., $\varepsilon=1$) is arbitrary and fails to account for the unique context of each application. 
    An overly conservative privacy level may render a model useless for its intended purpose (e.g., failing to detect a medical condition), while an overly permissive one can lead to privacy breaches, eroding user trust and incurring regulatory penalties~\cite{dwork2019differential,nanayakkara2022visualizing}.
    Therefore, a principled methodology is needed to help a decision-maker navigate this trade-off and select a configuration that balances privacy needs with performance requirements for deploying the differentially private models.

    \begin{figure}[t]
    \begin{center}
    \centerline{\includegraphics[width=1.0\columnwidth]{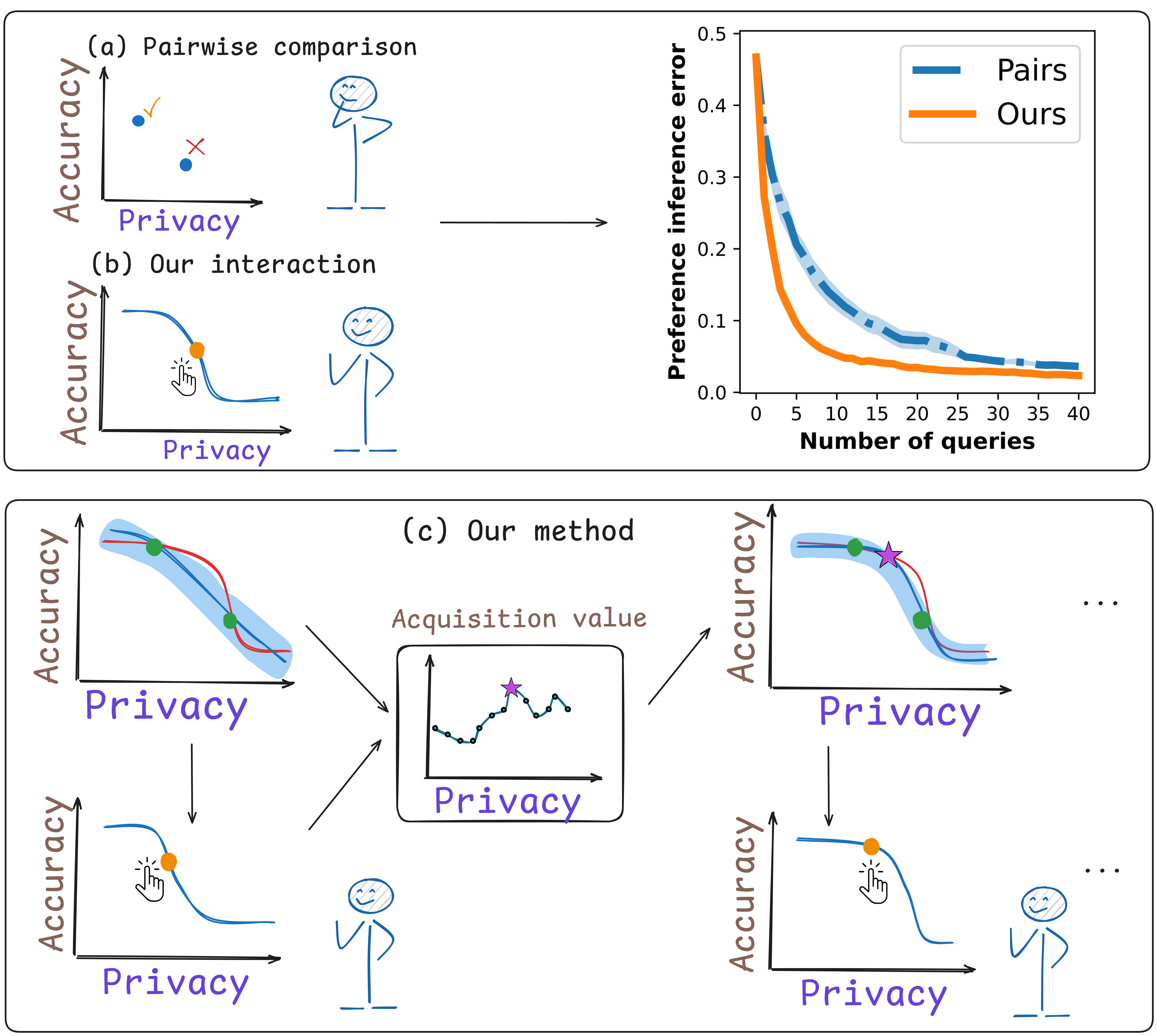}}
    \caption{
    An overview of our interactive framework for finding the optimal privacy-utility trade-off. (Top Panel) Unlike standard pairwise comparisons (a), our interaction (b) elicits richer feedback by asking the user to select their ideal point on a hypothetical curve. This leads to faster convergence in learning user preferences. (Bottom Panel) Our method alternates between updating the user preference learning model based on their choice (orange point) on the most informative curve (blue line) and updating the model of the true Pareto front (red line) based on an acquisition function to select the next privacy level to evaluate (purple star).}
    \label{fig:framework}
    \end{center}
    \vskip -0.2in
    \end{figure}

    The canonical solution to this trade-off is multi-objective optimization (MOO)~\cite{branke2008multiobjective}. 
    The traditional approach in MOO is to find the entire Pareto front -- the full set of all optimal trade-offs~\cite{branke2008multiobjective, gunantara2018review, avent2020automatic}. 
    However, this approach is often impractical for two reasons: it is \textbf{computationally infeasible}, requiring an exhaustive search of configurations that is too expensive for large models, and \textbf{practically unnecessary}, as a decision-maker typically only needs one final configuration for deployment~\cite{ozaki2024multi,lin2022preference}.

    A more practical path is \emph{interactive} MOO, which concurrently explores the trade-off space and learns the decision-maker's preferences. 
    This allows it to converge on the single optimal solution that aligns with their needs, without the unnecessary cost of recovering the entire Pareto front.
    While the standard interactive method is well-studied in non-DP settings~\cite{astudillo2020multi,ozaki2024multi,giovanelli2024interactive,lin2022preference}, it functions as a generic, black-box approach, making it inefficient when applied to DP, as it fails to exploit the unique underlying structure and suffers from two major drawbacks.
    First, it explores the entire high-dimensional hyperparameter space, building complex surrogate models for the objectives (privacy and accuracy in our case) to locate the Pareto front. 
    Second, it typically relies on simple pairwise comparisons (``is A better than B?'')~\cite{chu2005preference}, which are information-poor and place a high cognitive burden on the decision-maker.

    In contrast, our approach leverages the key property of DP -- optimizing accuracy at a fixed privacy level in DP naturally generates a solution on the Pareto front.
    Building on this, one of our contributions is to provide the theoretical grounding for the empirically observed S-shape of this trade-off. 
    This new theoretical insight allows us to directly and efficiently model the entire Pareto front as a sigmoid function after observing only a few of its points.

    We exploit these findings by proposing a more efficient interactive framework specifically tailored to navigate the privacy-accuracy trade-off in DP (\Cref{fig:framework}).
    Concretely, we estimate the Pareto front as a function that maps the privacy level to the best possible accuracy.
    This ``best'' accuracy is determined by an extensive HPO process, ensuring each point represents a high-performing configuration.
    We propose a Bayesian inference technique to build surrogates for the Pareto front. 
    Then we replace the pairwise comparisons commonly used in preference learning with a new interaction scheme where, the decision-maker selects their most preferred point directly on hypothetical trade-off curves. 
    This provides richer feedback per query by revealing the decision-maker's preferred trade-off along a continuous curve, rather than just the simple binary preference learned from a pairwise comparison, reducing the number of interactions needed to identify the optimal solution.

    Altogether we make the following contributions:
    \begin{enumerate}
        \item We propose a new interactive methodology that is significantly more efficient at both approximating the Pareto front and learning decision-makers' preferences in DP than existing general-purpose methods.
        \item We provide the first, to our knowledge, practical and principled guidance that empowers decision-makers to identify their subjectively optimal privacy level based on the tradeoff between privacy and accuracy, moving the choice of the optimal privacy level from a vague ``social problem'' to a structured, data-driven process.
        \item We demonstrate that our method is consistently more sample efficient in both learning the preferences and finding the optimal privacy level across multiple different private machine learning tasks than existing state-of-the-art approaches. 
    \end{enumerate}

\section{Background} \label{sec:bg}

    We begin with differential privacy in \Cref{subsec:dp}, which establishes the fundamental privacy-accuracy trade-off we aim to solve. 
    We then introduce multi-objective optimization in \Cref{subsec:moo}, the mathematical lens through which we formalize the trade-off in differential privacy.
    Finally, we cover interactive multi-objective optimization in \Cref{subsec:interactiveMOO}, the specific paradigm for solving such problems efficiently with a human-in-the-loop, which forms the basis of our contribution.

\subsection{Differential Privacy} \label{subsec:dp}

    There is a fundamental trade-off in DP between privacy guarantee (e.g., privacy budget $\varepsilon$ in $(\varepsilon, \delta)$-DP~\cite{dwork2006calibrating}, $\rho$ in $\rho$-zCDP~\cite{bun2016concentrated, dwork2016concentrated} and $\mu$ in $\mu$-GDP~\cite{dong2022gaussian}) and the resulting model's performance (e.g., accuracy in classification tasks). 
    We take $(\varepsilon, \delta)$-DP for illustration.
     
    \begin{definition}[$(\varepsilon,\delta)$-Differential Privacy] \label{def:dp}
        A randomized mechanism $\mathcal{A}$ is $(\varepsilon,\delta)$-differentially private \citep{dwork2006calibrating} if for any two neighboring datasets ${D}\in\mathcal{D}$ and ${D}'\in\mathcal{D}$ differing in only one record (adding or removing), and for any $S \subseteq \mathcal{S}$ in the output space $\mathcal{S}$,
        \begin{equation}
            P[\mathcal{A}(D)\in S]\leq e^{\varepsilon} \times P[\mathcal{A}(D')\in S]+\delta.
        \end{equation}
    \end{definition}

    The parameter $\delta$ serves as a relaxation for low-probability events that the guarantee fails to hold and usually is set to be a sufficiently small number. 
    When $\delta=0$, $(\varepsilon, \delta)$-DP becomes pure $\varepsilon$-DP. 
    The parameter $\varepsilon$ is called the privacy budget. 
    It controls the strength of privacy guarantee provided in \Cref{def:dp}: a smaller $\varepsilon$ provides more protection by ensuring that the probability distributions of the mechanism's output on any two neighboring datasets are statistically closer. 
    This increased indistinguishability makes it more difficult for an adversary to infer the presence of any single individual in the dataset, thus enhancing privacy but degrading accuracy, creating an inevitable trade-off. 

    Though we take $(\varepsilon, \delta)$-DP for illustration, this trade-off also exists for other definitions of DP.
    Therefore, in this work, we generalize the privacy guarantee using one parameter, $p$, which we refer to as the \textbf{privacy level}.
    To frame the trade-off as a consistent maximization problem, we define our privacy level, $p$, such that \textbf{a higher value corresponds to a stronger privacy guarantee.} 
    This allows our framework to treat both privacy and accuracy as objectives to be maximized.
    For common DP definitions, this can be achieved through a simple transformation. 
    For instance, in the case of $(\varepsilon, \delta)$-DP, where a smaller $\varepsilon$ means stronger privacy, we define the privacy level $p=-\log(\varepsilon)$ in this work. 

    Differentially Private Stochastic Gradient Descent (DP-SGD) \cite{abadi2016deep, song2013stochastic, rajkumar2012differential} and Differentially Private Adam (DP-Adam) are two widely used privacy-preserving optimization algorithms, derived from the standard SGD \cite{robbins1951stochastic} and Adam \cite{kinga2015method} optimizers. 
    To enforce privacy, these private optimizers first {clip the per-sample gradients}, which limits the maximum influence of any single data point and bounds the sensitivity. 
    Afterward, they {add} carefully calibrated statistical {noise} to these clipped gradients, obscuring individual contributions. 
    While this clip-then-noise procedure enables private training, it introduces a complex trade-off between the privacy level and model accuracy.
    The noise required for privacy, which depends on the hyperparameters, typically reduces model accuracy, and the choice of training hyperparameters (e.g., learning rate, clipping threshold, batch size, number of epochs) for a fixed privacy level further determines the final performance.
    Formally, we define our two objectives: privacy $f_p: \bs{\Theta}\rightarrow P$, which maps a set of hyperparameter to its privacy level, and accuracy $f_{\alpha}:\bs{\Theta}\rightarrow A$, which maps a set of hyperparameter to the resulting model accuracy~\cite{avent2020automatic}. 

    To achieve the best possible model performance for a given dataset $D$, a DP algorithm $\mathcal{A}$, and a privacy level $p\in P$, practitioners must carefully tune the training hyperparameters, $\bs{\theta}$, from the search space $\bs{\Theta}$. 
    This process, known as Hyperparameter Optimization (HPO), can be formalized as a constrained optimization problem in DP. 
    The goal is to find the optimal set of training hyperparameters that satisfies the privacy constraint and maximizes the model's accuracy:
    \newcommand{\AlgPerf}{f_{\text{acc}}}
    \begin{equation}
        \label{equ:hpo}
        h(p)=\max_{\bs{\theta}}f_{\alpha}(\bs{\theta};\mathcal{A},D) \; s.t.\; f_{p}(\bs{\theta}; \mathcal{A},D)\geq p.
    \end{equation}
    Thus, $h(p)$ is a mapping from the privacy level $p$ to the corresponding (best) accuracy obtained by HPO process.
    
    If the privacy level $p$ changes for the given task, the training hyperparameters also need  be re-optimized to maximize the model's accuracy $f_{\alpha}$ under the new privacy constraints.

    A significant body of work has focused on the HPO process in DP.
    For instance, one can transfer hyperparameters tuned on a smaller amount of data~\citep{koskela2023practical, sander2023tan}, extrapolate from hyperparameters tuned at smaller $\varepsilon$~\citep{pandanew}, perform fully differentially private hyperparameter tuning~\citep{liu2014private, papernot2022hyperparameter}, or even train using methods that eliminate the need for hyperparameter tuning altogether~\citep{bu2025towards}.

\subsection{Multi-Objective Optimization} \label{subsec:moo}

    The trade-off in DP between privacy and accuracy can be formulated as a multi-objective optimization (MOO) problem, which has multiple objectives $\bs{f}=(f_1,\dots,f_s)$ with $f_i:\bs{X}\rightarrow Y_i$:

    \begin{equation} \label{eq:moo}
        \arg\max_{\bs{x}} \{f_1(\bs{x}),f_2(\bs{x}),\dots,f_s(\bs{x})\} \text{ subject to } \bs{x} \in \bs{X}.
    \end{equation}    
       
    As noted, in DP the hyperparameters are the input space $X := \Theta$ and privacy $f_p: \Theta \rightarrow P$ and accuracy $f_\alpha: \Theta \rightarrow A$ are the objectives.

    The core concepts in MOO are the \emph{Pareto front} and \emph{dominating}.
    A candidate $\bs{x}$ \emph{dominates} another $\bs{x}'$ if its objective value is higher on all objective functions. 
    Typically, there is no single solution that dominates all others and, instead, there is a \emph{Pareto front} of all non-dominated solutions~\cite{branke2008multiobjective}.
    
   \paragraph{Pareto Optimality} For a pair $(\bs{x}, \bs{x}')$,  we say ``$\bs{x}$ weakly dominates $\bs{x}'$'' if $\bs{x}$ is no worse than $\bs{x}'$ in all objectives, which means $f_i(\bs{x}) \geq f_i(\bs{x}')$ for all $i\in \{1,\dots,s\}$. 
    If at least one of the inequalities is strict, we say ``$\bs{x}$ dominates $\bs{x}'$''. 
    If $\bs{x}$ is not (weakly) dominated by any other $\bs{x}'$ in the domain, $\bs{x}$ is called (weakly) Pareto-optimal. 
    \emph{(Weak) Pareto front} is a set of (weakly) Pareto optimal points.

     \paragraph{Multi-objective Bayesian Optimization (MOBO)} There are various methods to locate the Pareto front -- one of them is MOBO, which is designed for problems where the objective functions are expensive, black-box functions.
     Instead of optimizing the true objectives directly, MOBO builds a cheap-to-evaluate surrogate model (usually a Gaussian Process~\cite{williams1995gaussian}) for each objective function. 
     It then uses an acquisition function (like Expected Hypervolume Improvement~\cite{daulton2023hypervolume,yang2019multi} or based on information theory~\cite{belakaria2019max,suzuki2020multi}) to decide which point is the most promising to evaluate next with the real, expensive function. 
     \citet{avent2020automatic} employs MOBO to learn the trade-off between $\varepsilon$ and classification error in DP.

    \paragraph{Bayesian Inference}
    While most machine learning is interested in finding some maximally likely solution (model), Bayesian inference provides a full distribution over the quantities of interest~\cite{gelman1995bayesian}.
    This, among other things but most importantly to MOBO, provides uncertainty estimates.
    In Bayesian inference, we assume a prior (in this case $p(\bs{f})$) and apply the Bayes-rule to infer the posterior given some data $\{(x_j, \bs{y}_j)\}$:
    
    \begin{equation}\label{eq:bayes-posterior}
        p(\bs{f} \mid \{x_j, \bs{y}_j\}) \propto p(f) \prod_{i,j} p(y_{i,j} \mid f_i, x_j).
    \end{equation}

    Typically, estimating this posterior is challenging, but in the case that the likelihood  is Gaussian, then there is a closed-form solution in the form of Gaussian Processes~\cite{williams1995gaussian}.
    Hence, most literature (and our work too) makes the assumption that the observations --- the objective functions --- have Gaussian noise.

    \paragraph{$c$-Constraint Method}\footnote{The terminology ``$\varepsilon$-constraint method'' is usually used in MOO, but here it would become confused with the privacy budget $\varepsilon$. 
    Hence we call the approach here ``$c$-constraint method''.}
    Besides MOBO, another straightforward approach to finding the Pareto front is to select one of the objective functions to be optimized, and convert the others into constraints~\cite{branke2008multiobjective}, when not all objectives are expensive functions:
    \begin{equation} \label{eq:constraint}
        \max \; f_{l}(\bs{x}) \; s.t.\; f_i(\bs{x})\geq c_i, \; i=1,\dots, s,\; i\neq l.
    \end{equation} 
    The solution of~\Cref{eq:constraint} can be proven always to be weakly Pareto optimal~\cite{branke2008multiobjective, mavrotas2009effective, pirouz2016computational}.
    If there is a unique solution to ~\Cref{eq:constraint}, it can be proven to be Pareto optimal.  
    In DP, the privacy level naturally serves as the constraint.

\subsection{Interactive MOO: Preference Learning} \label{subsec:interactiveMOO} 

    Because Pareto optimal solutions have no objective ranking, ultimately a decision-maker needs to pick the preferred one.
    In fact, it is inefficient to first explore the whole Pareto front to only pick one at the end.
    Instead, \emph{interactive MOO} aims to learn the decision-makers' preferences \emph{while} simultaneously exploring the Pareto front, in order to find the best solution as efficiently as possible.
    This task has two components: (i) learning the objective functions $\bs{f}$ and (ii) learning the preferences {over the objectives $\bs{y}=(y_1,\dots,y_s)$, where $y_i = f_i(\bs{x})$.}
    We already discussed the first in~\cref{subsec:moo}, and will consider the second below.

    Let $U(\bs{y}; \bs{w})$ be a utility function that quantifies a decision-marker's preference over objectives $\bs{y}$, parameterized by the preference weights $\bs{w}$.

    To find the point that maximizes this utility\footnote{Whereas utility is traditionally measured only by model performance (e.g., accuracy) in DP literature, we define utility jointly over privacy and accuracy.}, a scalarization function is used to convert the vector of multiple objectives into a single scalar value.
    Chebyshev scalarization function~\cite{ozaki2024multi,ungredda2023elicit} is widely used in MOO as the utility function:
    \begin{align} 
    \label{eq:cheby_u}
        U(\bs{y}; \bs{w})& := \min \left (\frac{y_1}{w_1}, \dots, \frac{y_s}{w_s}\right),\\
        & \text{where } \sum_{i=1}^s w_i=1 \; \text{and}\; y_i=f_i(\bs{x}).\notag
    \end{align}    
    where both $\bs{f}$ and $\bs{w}$ are unknown and need to be learned.

    \Cref{eq:cheby_u} can identify any point on the Pareto front by varying the preference weights.
    Therefore, a decision-maker's preference for any specific optimal trade-off can be represented.
 
    Decision-makers are usually unable to mathematically describe the parameters of this function, or report the utility $U(\bs{y}; \bs{w})$ of a set of objectives~\cite{chu2005preference, furnkranz2010preference}.
    Instead, the most common technique for preference learning is to ask the decision-maker to pick their preferred solution among $q$ options $\{\bs{y}_j\}_{j=1}^q$~\cite{de2024preference, astudillo2023qeubo}.
    Then, given a \emph{user model} of how people make such choices, we can learn the underlying function $U$ from items that the decision-maker chooses~\cite{gonzalez2017preferential,astudillo2023qeubo,de2024preference,furnkranz2010preference}. 
 
    The most widely used model--and the one adopted in our work--is the Boltzmann-rational model~\cite{jeon2020reward,yamagata2024relatively}, which says the probability of choosing an item $\bs{y}^c$ among a set of candidates $\{\bs{y}_j\}_{j=1}^q$, given the preferences weight $\bs{w}$, is:
    \begin{equation}\label{eq-user_model}
       p(\bs{y}^c \mid \{\bs{y}_j\}_{j=1}^q, \bs{w}) = \frac{ \exp(U(\bs{y}^c ; \bs{w})/T) } { \sum_j^q \exp(U(\bs{y}_j;\bs{w})/T) },
    \end{equation}
    where $T$ is the temperature coefficient, controlling the rationality. 
    A lower $T$ means higher rationality, leading to a more deterministic choice, while a higher $T$ means lower rationality, leading to a more random and noisy choice. 

    \Cref{eq-user_model} is introduced as a statistical model for the likelihood in a Bayesian approach.
    In particular, given a dataset $\{ \bs{y}^*, \{\bs{y}_j\}_{j=1}^q \}$ of the preferred solution $\bs{y}^*$ among candidates $\{\bs{y}_j\}_{j=1}^q$, the posterior over preferences follows from the Bayes-rule given a prior $p(\bs{w})$:
    \begin{equation}\label{eq:posterior_preference}
       p(\bs{w} \mid \{ \bs{y}^*, \{\bs{y}_j\}_{j=1}^q \}) \propto  p(\bs{y}^* \mid \{ \bs{y}_j \}_{j=1}^q, \bs{w}) \times p(\bs{w}).
    \end{equation} 

    The next thing, then, is to decide which pairs of trade-offs to query the decision-makers with, for which we use Knowledge Gradient (KG)~\cite{frazier2009knowledge} -- a one-step Bayes-optimal acquisition function.
    It compute the expected improvement in the maximum utility gained from evaluating a pairs of trade-off $(\bs{y}_a, \bs{y}_b)$ when $q=2$:
    \begin{equation}
    \label{eq: kg}
        \text{KG}_t(\bs{y}_a,\bs{y}_b):= \Exp_{\bs{\beta}}[U_{t+1}^*-U_t^* \mid \text{query}=(\bs{y}_a, \bs{y}_b)],
    \end{equation}
    where $U^*_t$ is the maximum expected utility before the query, and the expectation is taken over the possible outcomes $\bs{\beta}$ of the user's choice for that pair $(\bs{y}_a, \bs{y}_b)$.
    
 \section{Problem Statement} \label{sec:research_problem}

    We operate under the standard ``\textit{trusted curator}'' threat model that is common for practical deployments of DP~\cite{avent2020automatic}. 
    In this model, a trusted internal team has legitimate access to the sensitive data and is responsible for the entire model development and selection process. 
    The adversary is an external party who only observes the single, final model that is ultimately deployed.
    In this paper, the terms ``\textit{user}'' and ``\textit{decision-maker}'' refer specifically to practitioners (e.g., model developers), not the individuals contributing sensitive data. Our objective is to guide these practitioners in selecting an optimal privacy level $p$ for deploying the differentially private models.
    
    With these assumptions, we formulate the task of balancing privacy and accuracy in DP as a MOO problem. 
    Formally, we are interested in finding the set of hyperparameters $\theta \in \Theta$ that maximize the accuracy $f_{\alpha}: \Theta \rightarrow A$ and privacy $f_{p}: \Theta \rightarrow P$.

    Because these two objectives are in conflict, no single solution can simultaneously maximize both. 
    For example, improving the privacy will result in a sacrifice in the accuracy. 
    Instead, there is the Pareto front containing the set of all optimal trade-offs where one objective cannot be improved without degrading the other.
    The goal of interactive MOO is to find the preferred trade-off on the Pareto front that best aligns with the decision-maker's preference.

    We assume the decision-maker has a latent (unknown) utility function, $U(p, \alpha;\bs{w})$, which quantifies the overall satisfaction with a given privacy-accuracy trade-off $(p, \alpha)$. 
    Then the goal turns to find the point on the Pareto front that maximizes this utility function.
    Therefore, our central research problem is to design an efficient, interactive framework that solves the following optimization problem to find the preferred trade-off:
    \begin{equation}
        \label{equ:optimize_u}
        \max_{\bs{\theta}} \;U(f_p(\bs{\theta}), f_\alpha(\bs{\theta});\bs{w}).
    \end{equation}

\section{Methodology} \label{sec:method}

    To solve \Cref{equ:optimize_u}, we propose an interactive MOO framework that is specifically tailored to the structure of DP, making it more efficient than standard approaches.
    Standard interactive methods build generic surrogate models for each objective (privacy and accuracy in our case) and learn user preferences from simple pairwise comparisons.
    This black-box approach is inefficient as it fails to leverage key properties of DP: the privacy level itself can be used as a direct constraint. 
    By fixing a privacy level and optimizing for the best possible accuracy via HPO, we can directly generate a solution on the Pareto front. 
    The maximization operation $h(p)$, defined in \Cref{equ:hpo}, returns the best possible accuracy achievable for a given privacy level $p$. 
    Therefore, the curve of this function, $(p,h(p))$, constitutes the Pareto front: the set of all optimal, non-dominated trade-offs.
    We show our motivation of taking S-shaped functions as surrogates to model Pareto fronts theoretically and empirically below, which allows us to model the entire front directly and efficiently with a Bayesian sigmoid function.    
    
    First in \Cref{subsec:pareto_front}, we introduce our motivation of Gompertz and sigmoid priors and our Bayesian model of the Pareto front and how we use the privacy level constraint optimization to gather the data necessary for inference.
    In~\Cref{sec:pareto-preferences}, we discuss how to model the decision-maker feedback when querying on hypothetical Pareto fronts.
    \Cref{sec:pareto-acq} describes how we are able to solve both the tasks of preference and Pareto front learning in a single decision, based on a knowledge-gradient acquisition function. 
    Lastly, the whole framework is summarized in~\Cref{subsec:algo}.
 
\subsection{Pareto Front Modeling}\label{subsec:pareto_front}
    The key insight driving this work is that the Pareto front $h: P \rightarrow A$ can be modeled by analytical S-shaped functions in general.
    Here we provide theoretical support and empirical evidence for this insight.

    \paragraph{\textbf{Trade-off between privacy and accuracy of logistic regression}}  
    We start with pure $\varepsilon$-DP and use a well-understood mechanism--output perturbation for regularized logistic regression~\cite{chaudhuri2008privacy}--as a concrete case study to generate and model this tradeoff curve between $\varepsilon$ and accuracy. 
    Following the Algorithm 1 in~\cite{chaudhuri2008privacy}, let $(x_1,y_1),\dots,(x_n,y_n)$ be a set of labeled points over $\mathbb{R}$ with $x_i$ uniformly sampled from $[-1,1]$ for all $i$. 
    First, we compute the coefficient $\xi$ obtained by the logistic regression with a regularization constant $\lambda$. 
    Adding noise $\eta$ to $\xi$ gives the noisy coefficient $\xi_{noisy}=\xi + \eta$ with 
    $\eta \sim \text{Lap}(S_f/\varepsilon)$. $S_f$ is the sensitivity of function $f$, which is at most $\frac{2}{n\lambda}$ for logistic regression~\cite{chaudhuri2008privacy}. 

    Here we derive our theoretical formulation of {$h(\varepsilon)$}.
    The probability of correct prediction for a positive class is 
    \begin{align}
        P((\xi+\eta)x\geq 0)&=P(\eta x \geq -\xi x)\\  \notag
        &= P(Z\geq- \xi x)\quad \text{where } Z=\eta x\\ \notag
        &=1-F_Z(-\xi x)\\ \notag
        &=1-0.5\cdot \exp(-\frac{|\xi|}{S_f}\cdot \varepsilon). \notag
    \end{align} 
    The last equation is due to $Z\sim \text{Lap}((S_f/\varepsilon)|x|)$. 
    The derivation is similar for a negative class:
    \begin{align}
        P((\xi+\eta)x\leq 0)&=P(\eta x \leq -\xi x)\\  \notag
        &=P(Z\leq- \xi x)\quad \text{where } Z=\eta x\\ \notag 
        &=F_Z(-\xi x)\\ \notag 
        &=1-0.5\cdot \exp(-\frac{|\xi|}{S_f}\cdot \varepsilon). \notag
    \end{align} 
    Thus the theoretical accuracy for any single point $x$ is 
    \begin{align}
        h(\varepsilon, x)=1-0.5\cdot \exp(-C\varepsilon),
    \end{align}
    where $C=\frac{|\xi|}{S_f}$. 
    The expectation accuracy over the entire distribution of the data is 
    \begin{align}
    \mathbb{E}[h(\varepsilon)]&=\int_{-1}^1h(\varepsilon, x)\cdot p(x) dx\\ \notag
    &= h(\varepsilon) \cdot \int_{-1}^1 p(x) dx \\ \notag
    &=1-0.5\cdot \exp(-C\varepsilon)\\ \notag
    &=1-0.5\cdot \exp(-C\exp(\tilde{\varepsilon})), \notag
    \end{align}
    where $\tilde{\varepsilon}=\log(\varepsilon)$. This is a special sigmoid curve - Gompertz function~\cite{winsor1932gompertz}. 
    Unlike the standard sigmoid, the Gompertz function is asymmetric: the curve approaches its right-side asymptote much more gradually than its left-side asymptote.

    Next we collect empirical data on the model's performance across a range of privacy budgets $\varepsilon\in [0.0001, 0.15]$ to approximate the Pareto front. 
    Our theoretical derivation for differentially private logistic regression proves that its Pareto front follows a Gompertz function. 
    The Gompertz curve is a member of the broader family of sigmoid (S-shaped) functions, all of which are characterized by a slow initial growth, a rapid increase, and a final saturation towards a plateau.
    Based on this intuition, we also model the observed S-shaped relationship with a more general and widely-used functional form - the sigmoid function with parameters $\bs{\beta}=(L,b, k,c)$ to model the trade-off curve:
    \begin{equation} 
        \label{eq:sig}
        h({\varepsilon}; L, k, b, c) = \frac{L}{1+\exp(-k(\log{\varepsilon}-c))} + b,
    \end{equation}
    where $b$ is the model's minimum accuracy and $L+b$ is the model's maximum accuracy. 
    $k$ is the steepness of the curve and $c$ is the infection point, at which the accuracy is exactly halfway.
    By taking $p=-\log(\varepsilon)$, we get
     \begin{equation} 
        \label{eq:sig-p}
        h(p; L, k, b, c) = \frac{L}{1+\exp(k(p-c))} + b.
    \end{equation}

    \begin{figure}[t]
    \begin{center}
    \centerline{\includegraphics[width=1.0\columnwidth]{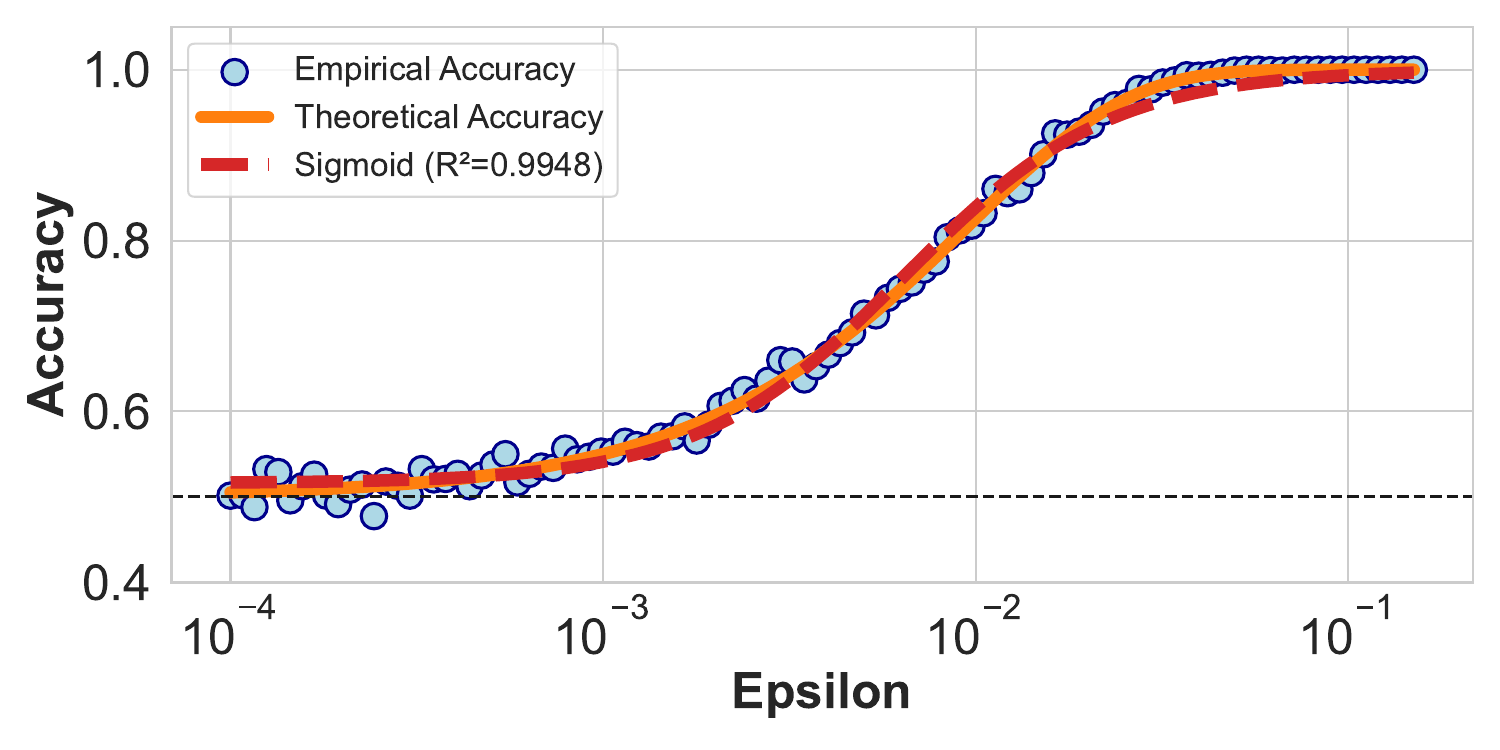}}
    \caption{The theoretical (Gompertz) and empirically fitted (sigmoid) privacy-accuracy tradeoff for differentially private logistic regression. The sigmoid curve presents a good fit to theoretical accuracy.}
    \label{fig:sigmoid}
    \end{center}
    \end{figure}

     We use \texttt{scipy.optimize.curve\_fit} in Python to fit the data.
     \Cref{fig:sigmoid} shows the sigmoid fitted results and indicates that the sigmoid function provides a good approximation of the Pareto front in logistic regression. 
    We then generalize this observation to a broader class of differentially private models.
    
    \paragraph{\textbf{Pareto front modeling in DP models}}    
    
    While this S-shaped relationship was first observed in the context of $\varepsilon$-DP, we posit that this trend is not specific to this definition but is a fundamental characteristic of the privacy-accuracy trade-off. 
    The rationale for this generalization is that the shape of the Pareto front is governed by model-agnostic principles. 
    For any DP model, the tradeoff is bounded by two asymptotes: a lower performance bound at a higher privacy level (e.g., $\varepsilon\rightarrow 0$), where high noise dominates, and an upper bound at a lower privacy level (e.g., $\varepsilon\rightarrow\infty$), where the model's accuracy approaches its non-private counterpart. 
    The transition between these bounds inherently follows a pattern of diminishing returns, where initial decreases in the privacy level yield substantial accuracy gains that taper off as the model's performance saturates. 
     \begin{figure}[t]
    \begin{center}
    \centerline{\includegraphics[width=1.0\columnwidth]{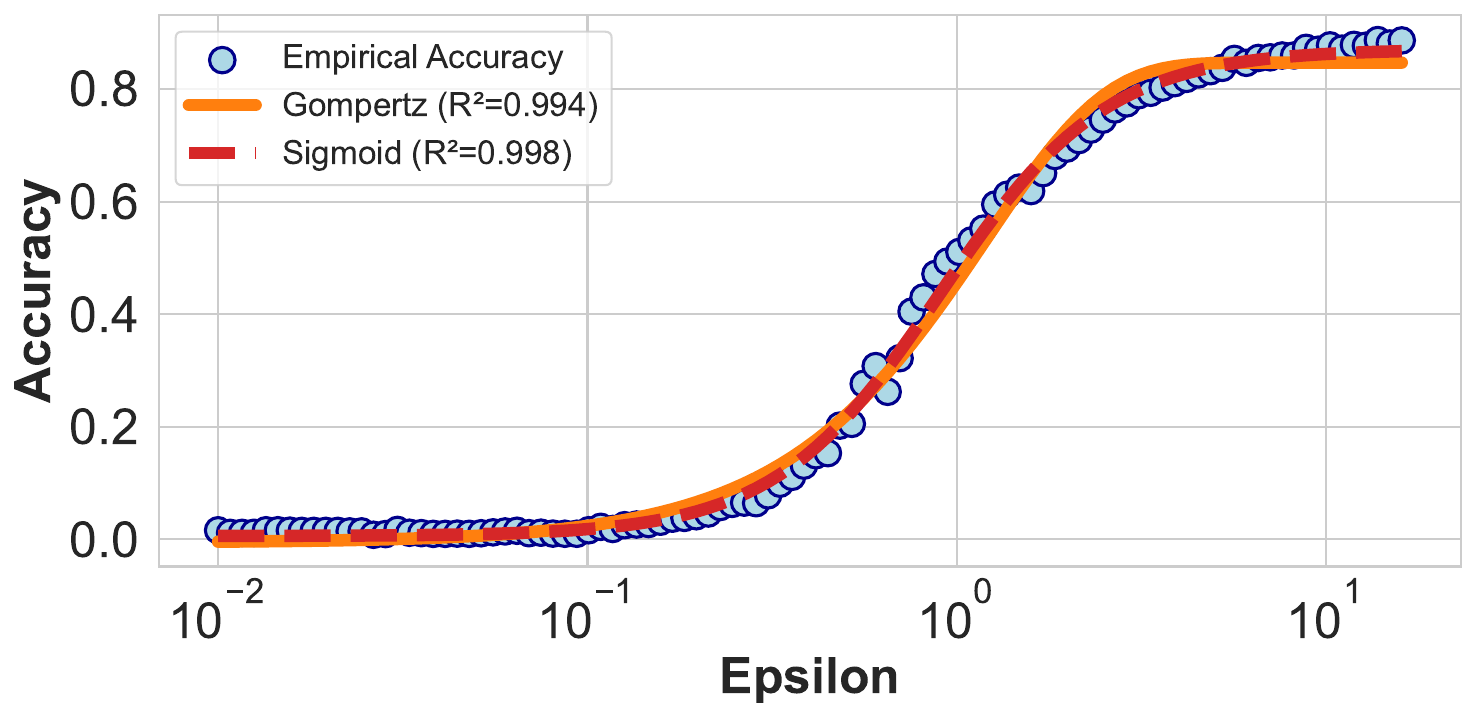}}
    \caption{The Gompertz and sigmoid curves fitting to privacy-accuracy tradeoff for differentially private transfer learning on CIFAR100 with $\delta=10^{-5}$. Both curves effectively capture the trend of the observed trade-off data.}
    \label{fig:cifar100-sigmoid}
    \end{center}
    \end{figure}

    In \Cref{fig:cifar100-sigmoid}, we present the trade-off data after performing HPO for each privacy level of differentially private transfer learning on CIFAR100. 
    The details of HPO and the model can be found in \Cref{subsec:dp_results}. 
    Then we use both Gompertz and sigmoid to fit the trade-off data, which gives comparably good fitting results. 
    Given that this behavior is fundamental in DP, we believe that the S-shaped functions serve as general and effective models for the privacy-accuracy Pareto front across a wide range of models and datasets.

    Based on our theoretical and empirical findings, we propose to model the Pareto front $h_{\bs{\beta}}: P \rightarrow A$, parameterized by $\bs{\beta}$, between privacy and accuracy in general differentially private models as an S-shaped function.  
    In practice, the accuracy of any single model is a random variable due to stochastic elements in the training process and the noise inherent to the differential privacy mechanism. 

    To account for this variability, we adopt a probabilistic approach. 
    In particular, we assume noisy observations are drawn
    from a Gaussian distribution whose mean is the expected accuracy predicted by the S-curve.
    This gives us the following likelihood: $\alpha \sim \mathcal{N}({h}({p}; \bs{\beta})), \sigma^2)$, where $\sigma^2$ is the variance of the observation noise.
     Then, assuming a prior over the parameters ($p(\bs{\beta})$, $p(\bs{\sigma})$) and observed (maximum) accuracies given observed data points $\{{p}_n, \alpha_n\}_{n=1}^N$, standard Bayes-rule gives us the posterior: 
    \begin{equation}\label{eq:posterior_sigmoid}
        p(\bs{\beta}, \sigma  \mid   \{ {p}_n, \alpha_n \}_{n=1}^N ) \propto
        p(\bs{\beta}) p(\sigma) \prod_n^N \mathcal{N}(\alpha_n  \mid  {h}({p}_n); \bs{\beta}, \sigma^2). 
    \end{equation}
    
    We employ two type of S-shaped functions to model the Pareto front $h$ between the privacy level $p$ and the accuracy $\alpha$:

    \begin{enumerate}
        \item A Gompertz function parameterized by ${\bs{\beta}}=(L,k,b,c)$: 
        \begin{equation}\label{eq:gompertz}
           h^{g}({p};L,k,b,c)=\frac{-L}{\exp(-k \exp(-c{p}))} + b. 
        \end{equation}
         \item A sigmoid function parameterized by ${\bs{\beta}}=(L,k,b,c)$: 
        \begin{equation}\label{eq:sigmoid}
            h^{s}({p};L,k,b,c)=\frac{L}{1+\exp(k({p}-c))} + b.    
        \end{equation}
    \end{enumerate}

    Each point on our empirical trade-off curve is the result of a full HPO process, conducted for a fixed privacy level, $p$. 
    This ensures that every observed data point represents a Pareto-optimal solution for its specific privacy level.

\subsection{Interactive Preference Elicitation} \label{sec:pareto-preferences}

    Having established a method to construct a continuous surrogate model of the privacy-accuracy Pareto front, $h_{\bs{\beta}}$, we now address the subsequent challenge: eliciting a decision-maker's preferences to identify their optimal operating point. 
    As opposed to querying pairwise comparisons --- the current state-of-the-art --- we propose to exploit the fact that we have a model of the Pareto front to gather more informative feedback.
    In particular, we present \emph{hypothetical} Pareto fronts to the decision-maker, parameterized by $\bs{\beta}$, and ask them to pick their favorite trade-off $\bs{y}^* = {(p, \alpha)}$ on it to elicit the most informative feedback about their preferences.

    This proposal naturally leads to the central modeling question: \textbf{how do we formally interpret the decision-maker's choice?}
    We follow the typical modeling assumption that the user has some latent utility function to quantify the quality of a trade-off $U(\bs{y};\bs{w})$, and that the likelihood of picking a trade-off is proportional to this utility value.
    We argue, in this setting, that \emph{the decision-maker interprets the presented front as a (technically infinite) set of trade-offs}.
    We model this by extending the Boltzmann-rational model (recall~\cref{subsec:interactiveMOO}):

    \begin{equation}
        p(\bs{y^*} \mid \bs{\beta}, \bs{w}) = \frac{ \exp(U(\bs{y}^* ; \bs{w})/T) } { \int_{\bs{y}\in h_{\bs{\beta}}} \exp(U(\bs{y}_i;\bs{w})/T) d\bs{y}}.
    \end{equation}    
    For practical implementation, we discretize the continuous front into a dense set of $q$ points $\{\bs{y}_j\}_{j=1}^q$, transforming the integral in the denominator into a summation:
    \begin{equation} \label{eq:user_model}
        p(\bs{y^*} \mid \bs{\beta}, \bs{w}) = \frac{ \exp(U(\bs{y}^* ; \bs{w})/T) } { \sum_j^q \exp(U(\bs{y}_j;\bs{w})/T) }.
    \end{equation}   
    With this, we have a user (likelihood) model to interpret the decision-maker's trade-off choices.
    Now, learning the decision-maker's preference is the task of Bayes inference given a prior $p(w)$ and preference data $\{\bs{y}^*_m, \bs{\beta}_m\}_{m=1}^M$.
    \begin{equation}\label{eq:posterior_preference_curve}
       p(\bs{w} \mid \{ \bs{y}^*_m, \bs{\beta}_m\})_{m=1}^M \propto  p(\bs{w})\prod_m^M p(\bs{y}^*_m \mid \bs{\beta}_m, \bs{w}).
    \end{equation}

\subsection{Acquisition Functions}\label{sec:pareto-acq}
    
    So far, we have discussed Bayesian inference over the Pareto front (\Cref{subsec:pareto_front}) and decision-maker preferences (\Cref{sec:pareto-preferences}).
    This section introduces how to pick which (1) hypothetical Pareto fronts with parameters $\bs{\beta}$ to show to the decision-maker and (2) constraints $p$ to do (constrained) optimization on to find Pareto optimal trade-off $\bs{y} = {(p, \alpha)}$ that can help maximize the utility most.
    While typically this has been solved with two different acquisition functions on the two posteriors in interactive MOO methods, we adopt the approach by~\citet{ungredda2023elicit}, directly optimizing the knowledge-gradient on the utility.

    Knowledge-gradient~\cite{frazier2009knowledge} computes the expected utility under the posterior.
    When we have observed a set of $M$ preference observations $\{\bs{y}^*_m, \bs{\beta}_{m}\}_{m=1}^M$ and $N$ Pareto optimal observations $\{\bs{y}_n\}_{n=1}^N$, the expected utility of some trade-off $\bs{y}$ is computed by
    \begin{align}\label{eq:expected-utility-trade-off}
       &{\Exp_{\bs{\beta},\bs{w}}[U(\bs{y})]} = \\
       &\int_{\bs{\beta}, \bs{w}} U(\bs{y}; \bs{w}) p(\bs{w} \mid \{\bs{\beta}_{m},\bs{y}^*_m\}_{m=1}^M) p(\bs{\beta} \mid \{\bs{y}_n\}_{n=1}^N), \notag
    \end{align}
    where the posteriors $p(\bs{w} \mid \dots)$ and $p(\bs{\beta} \mid \dots)$ are given by \Cref{eq:posterior_sigmoid,eq:posterior_preference_curve}.
    The largest expected utility is then 
    \begin{equation}
    U^*_{M,N} := \max_{p}\Exp_{\bs{\beta},\bs{w}}[U(p, h_{\bs{\beta}}(p);\bs{w}]
    \end{equation}
    based on the current posteriors.

    Therefore, we maximize the KG with respect to the posterior over the preferences to find the curve to present to the decision-maker:
    \begin{align}\label{eq:curve}
         &\bs{\beta}_{M+1} \leftarrow 
         \arg \max_{\bs{\beta}} \Exp[U^*_{M+1,N}-U^*_{M,N} \mid \bs{\beta}_{M+1}=\bs{\beta}].
    \end{align}
    Similarly, when picking the constraint $p_{N+1}$ to optimize with, we maximize the KG with respect to the posterior over the Pareto front:
    \begin{align}\label{eq:epsilon}
         &p_{N+1} \leftarrow 
         \arg \max_{{p}} \Exp[{U^*_{M,N+1}-U^*_{M,N}} \mid {p}_{N+1}={p}].    
    \end{align}

\subsection{Algorithm}\label{subsec:algo}
    The proposed algorithm is summarized in Algorithm \ref{alg:interleave}.
    The final output is the estimated optimal trade-off that best aligns with the decision-maker's preferences. 
    
    \begin{algorithm}
       \caption{Interactive algorithm to identify the optimal trade-off.}
       \label{alg:interleave}
    \begin{algorithmic}
       \STATE {\bfseries Input:} Training dataset $\mathcal{D}$. DP-model $\mathcal{M}$.  Search range of hyperparameters $\Theta$. Priors of $\bs{w}$ and $\bs{\beta}$: $p(\bs{w})$ and $p(\bs{\beta})$, number of iterations NUM.
    
       \FOR{$t=1, \dots $NUM }
            \IF{Perform Preference Elicitation:}
                \STATE Select the next curve to present $\bs{\beta}_{M+1} \leftarrow$ Equation (\ref{eq:curve}).
                \STATE Observe decision-maker's choice $\bs{y}^*_{M+1}$ on front $\bs{\beta}_{M+1}$.
                \STATE Update posterior over preference $\bs{w}$ $\leftarrow$ Equation (\ref{eq:posterior_preference_curve}). 
            \ENDIF
            
            \IF{Perform Pareto Front Modeling:}
                \STATE Select the next privacy level ${p}_{N+1} \leftarrow$ Equation (\ref{eq:epsilon}).
                \STATE HPO for ${p}\geq p_{N+1}$ to get $\alpha_{N+1} \leftarrow$ Equation (\ref{equ:hpo}). 
                \STATE Update posterior on Pareto front $\bs{\beta} \leftarrow$ Equation (\ref{eq:posterior_sigmoid}).  
            \ENDIF
       \ENDFOR
       \STATE {\bfseries Output:} The optimal trade-off $(p^*, \alpha^*)$. 
    \end{algorithmic}
    \end{algorithm}    
    The computation of acquisition functions (Equations \ref{eq:curve} and \ref{eq:epsilon}) can be found in~\Cref{appendix:compute_acq}.
    For the experiments presented in this paper, we adopt a direct interleaving scheme, and sequentially perform one model evaluation to refine our estimate of the Pareto front, followed by one interaction to update our model of the decision-maker's preferences. 
    This ensures balanced and consistent progress on both learning objectives throughout the process.
    Our acquisition function framework also supports adaptive interleaving~\cite{ungredda2023elicit}: after evaluating the acquisition functions from both procedures, the algorithm compares their values to determine which action to take and updates the corresponding posterior distribution accordingly. 

\section{Related Work} \label{sec:related}

    \paragraph{Learning trade-off in DP}  
    Prior works address the trade-off between the privacy level and the model performance in DP using different multi-objective optimization methods.
    \citet{avent2020automatic} formulated the trade-off between $\varepsilon$ and classification error as multi-objective Bayesian optimization and use hypervolume as an index to perform HPO to get the entire Pareo front.
    \citet{priyanshu2022efficient} considered the linear combination of $\varepsilon$ and validation loss and employ three different methods -- Bayesian optimization, reinforcement learning and evolutionary approach for HPO. 
    However, \citet{priyanshu2022efficient} assumed the preference weights are known beforehand, leaving the critical question of how to choose them unanswered.

    The challenge of selecting an appropriate privacy level (e.g., $\varepsilon$) is a well-recognized open problem in the practical deployment of DP. 
    Foundational work highlights that the right choice of $\varepsilon$ is not universal but is highly context-dependent, varying between different applications and datasets~\cite{dwork2019differential}.
    To address this, one line of research focuses on making $\varepsilon$ more interpretable. These works aim to translate the abstract $\varepsilon$ value into more concrete and understandable measures of privacy risk, helping practitioners make a more informed judgment call~\cite{cummings2021need,nanayakkara2023chances}. 
    Another direction, more aligned with our own, focuses on building interactive systems. 
    \citet{nanayakkara2022visualizing} developed an interface to help users visualize the trade-offs between $\varepsilon$, accuracy, and disclosure risk. \citet{nanayakkara2024measure} proposed an interactive paradigm for exploratory data analysis which allows a user to start with a small $\varepsilon$ and iteratively increase it if better utility is required. 
    However, it still requires a predefined privacy budget, which is difficult to set without seeing the trade-off curve. 

    \paragraph{Interactive MOO}
    Incorporating the decision-makers' preferences into exploration algorithms has been studied in MOO literature.
    \citet{astudillo2020multi} proposed a multi-attribute BO algorithm with a preference-based expected improvement acquisition function.
    \citet{ozaki2024multi} proposed active learning for preference query in interactive MOO. 
    \citet{ungredda2023elicit} considered KG acquisition functions in both preference learning and objectives estimation procedures, and proposed a general framework on when to elicit preferences which we build our work on.
    However, these methods all use pairwise comparisons to collect decision-makers feedback. 
    While it is easy to implement, each comparison unfortunately provides only a small amount of preference information, and hence the approaches requires a considerable number of queries. 

     {\citet{ungredda2022single} proposed to solve this limitation by generating the estimated Pareto front using an evolutionary algorithm and presenting all discrete points on the Pareto font to the decision-maker.}
    However, due to the computational difficulty of generating the estimated front, they only perform a single round of interaction. 
     
    Practical applications of interactive MOO are attracting growing interest. \citet{giovanelli2024interactive} studied the trade-off in hyperparameter optimization between accuracy and power consumption. 
    They first generate the initial Pareto front and then interact with the decision-maker through pairs of points on the initial front.

\section{Experiments} \label{sec:experiments}
    \Cref{subsec:logistic_results} presents the experiments on finding the optimal trade-off of differentially private logistic regression.
    \Cref{subsec:dp_results} contains experiments on finding the optimal trade-off of differentially private deep transfer learning problems.  
    Then, in \Cref{subsec:ablation}, we present an ablation study on preference learning and another one on Pareto front modeling, to investigate sample efficiency in both tasks.  
    Lastly, we compare our utility function with other linear utility functions to justify our choice in \Cref{subsec:visulization_u}. 
    All results report the average and the standard error of $30$ runs.  

\subsection{Experiments Configurations}\label{subsec:exp_config}    
    \paragraph{Hyperparameter Optimization}

    For the inner-loop HPO task of finding the maximal accuracy for a given privacy level $p$ (Equation \ref{equ:hpo}), we require an efficient method for optimizing an expensive black-box function. 
    To this end, we select Bayesian Optimization (BO) due to its well-documented sample efficiency.
    The search range of hyperparameters can be found in \Cref{tab:hyper-bounds}.
    We employ a standard, non-private version of BO. 
    Under our standard ``trusted curator'' threat model, the intermediate results of our interactive optimization process--including all HPO evaluations and the generated Pareto front--are considered confidential to the internal team. 
    Therefore, we do not account for the cumulative privacy cost of the iterative search, as these exploratory steps are never exposed to the adversary. 
    The only privacy guarantee relevant to the threat model is that of the single, final model chosen for deployment, which has a well-defined privacy budget $p$. 

    However, it is crucial to note that our interactive framework is \textbf{agnostic to this choice of HPO methods.}
    One could easily substitute other methods, such as a more efficient HPO algorithm~\cite{bu2025towards}, without altering the core logic of our preference learning and Pareto front modeling. 
    This modularity is a significant feature, allowing our framework to be readily adapted to different computational constraints and development environments.

    \begin{table}[ht]
    \centering
    \begin{tabular}{@{}lll@{}}
    \toprule
    \textbf{Parameter}         & \textbf{LogReg+SGD}  &   \textbf{Fine-Tuning+Adam}           \\ \midrule
    \textbf{Batch size}        &   [8, 512]  & [192, $\lvert \mathcal{D} \rvert$]                   \\
    \textbf{Learning rate}    &     [5e-4, 5e-2]                  & [10e-5, 0.1] (log)          \\
    \textbf{Clipping threshold} &    [0.1, 4]                 & [10e-4, 100] (log)        \\ 
    \textbf{Epochs} & [1, 64]     & 40    
    \\ \bottomrule
    \end{tabular}
    \caption{Search bounds for hyperparameters during Bayesian optimization.}
    \label{tab:hyper-bounds}
    \end{table}
    
    \paragraph{Common Experimental Setup and Parameters}
    Here we list the common setup and parameters in all experiments.
    We split 10\% of the training set $\mathcal{D}$ to form a validation set and use the model's performance on this set as the optimization objective for hyperparameter tuning.
    We run 20 iterations of BO using the Optuna library~\cite{akiba_optuna_2019} with the BoTorch sampler that employs Gaussian processes. 
    Afterward, we conduct a final training run on the full $\mathcal{D}$ with the optimal hyperparameters found during tuning. 
    For a fair comparison, \textbf{we allow the baselines an equal number of 20 samples every step to estimate the objective functions}.

    We model the decision-maker's choices with Chebyshev utility functions and a Boltzmann-rational model with parameter $T=0.2$ (recall~\Cref{subsec:interactiveMOO,sec:pareto-preferences}).
    Our method exhibits robustness to the choice of $T$. We present the sensitivity analysis in \Cref{appendix:user-model}.
    The true (simulated) utility weights are sampled from a Dirichlet distribution parameterized by $(2, 2)$ (same as in ~\cite{ozaki2024multi}), which is also used as the prior for the solution methods. 

    \Cref{tab:priors} lists the prior distributions for parameters of sigmoid and Gompertz functions that we use in all experiments. 
    \begin{table}[ht]
    \centering
    \begin{tabular}{@{}lll@{}}
    \toprule
    \textbf{Parameter}         & \textbf{Sigmoid}           & \textbf{Gompertz}              \\ \midrule
    $L$      &  $\text{Beta}(40,2)$                & $\text{Uniform}(0.8,4)$  \\
    $k$      &  $\text{LogNormal}(\log(10), 0.2)$  & $\text{Uniform}(10,100)$  \\
    $c$      &  $\text{Beta}(2,2)$                 & $\text{Uniform}(1,10)$  \\
    $b$      &  $\text{Normal}(0, 0.1)$            & $\text{Uniform}(0.8,1.1)$  \\
    $\sigma$ &  $\text{Gamma}(0.5,0.1)$            & $\text{Gamma}(0.5,0.1)$  \\
    \bottomrule
    \end{tabular}
    \caption{Priors distribution for parameters of sigmoid and Gompertz functions.}
    \label{tab:priors}
    \end{table}

    We normalize both the accuracy $\alpha$ and privacy level, $p=-\log(\varepsilon)$, to the $[0,1]$ scale using a min-max transformation. 
    This process ensures that both objectives are comparable for preference learning, with 1 representing the most desirable outcome for each objective.
    This normalization is performed on a user pre-defined range of the privacy budget, $[\varepsilon_{\min}, \varepsilon_{\max}]$. 
    The discussion of the rationale and practical guidance for selecting this range are detailed in \Cref{sec:discussion}.
    Since this normalization is invertible, any point in the scaled space can be mapped directly back to its original $\varepsilon$ value.
    
    \paragraph{Evaluation Metrics}
    In total 20 steps, we adopt a direct interleaving scheme and sequentially perform one model evaluation at step $t$ and one interaction at step $t+1$.  
    We report error in \emph{preference inference} and \emph{regret} at step $t$.
    The error at step $t$ in preference estimation is the expected mean-squared error under the posterior over the preference weights $\bs{w}$ as is common (e.g.\cite{ozaki2024multi}):
    \begin{equation} \label{eq:pref-inf-err}
        \bs{w}_{\text{error}}^t = \Exp_{\bs{w}^t} \|\bs{w}_{\text{true}}-\bs{w}^t\|_2. 
    \end{equation}
 
    Regret at step $t$ is defined as the difference between the utility of the true optimal point, $\bs{y}^*_{\text{true}}$, and the (true) utility of the privacy level $p^t$ that maximizes the utility under the current posteriors:
    \begin{equation}
        {p}^t := \arg\max_{{p}} \Exp_{\bs{w}, \bs{\beta}}[U(({p},h_\beta({p}));\bs{w})].
    \end{equation}
    The regret at step $t$ is defined as 
    \begin{equation} \label{eq:regret}
        \text{Regret}^t =  U(\bs{y}_{\text{true}}^*; \bs{w}_{\text{true}})- U(({p}^t, h({p}^t)); \bs{w}_{\text{true}}).
    \end{equation}
    \paragraph{Baselines.}
    We compare our method with state-of-the-art baselines, all of which use GPs to model objective functions (privacy and accuracy in our case). 
    Therefore, we follow the method proposed in \cite{avent2020automatic} to build surrogates for privacy and accuracy. Details can be found in \Cref{appendix:baselines}.
    These baselines employ different acquisition functions for preference learning, but all of them use expected improvement (EI) to pick candidates in optimization stage. 
    1) \textbf{BALD}~\cite{ozaki2024multi} uses Bayesian Active Learning by Disagreement (BALD) to select which pairs to query the decision-maker.
    2) \textbf{UU}~\cite{astudillo2020multi} chooses queries pairs randomly based on the utility function.
    3) \textbf{Pareto}~\cite{ungredda2022single} also presents the Pareto fronts to the decision-maker for preference learning.
    The original work presents the Pareto front only once: they assume the decision-maker is perfect.
    We extend this method to be iterative --- our interactive setting --- and use our user model for inferring preferences, which allows the noisy feedback from decision makers.
    4) \textbf{qEUBO}~\cite{astudillo2023qeubo} chooses the expected utility of the best option (qEUBO), which is equivalent to KG acquisition function. We set $q=4$, as proposed in ~\cite{astudillo2023qeubo}, which means 4 points are picked every step.
    5) \textbf{TS}~\cite{gonzalez2017preferential} employs duelling-Thompson sampling (TS) as one of the dueling acquisition functions for preference learning. 
    6) \textbf{EPIG}~\cite{smith2023prediction} proposes the expected predictive information gain (EPIG), an acquisition function that measures information gain in the space of predictions rather than parameters.

    To demonstrate the effectiveness and broad applicability of our proposed interactive framework, we conduct experiments across two distinct and representative machine learning scenarios on six datasets.

\subsection{Finding the Optimal Trade-off in Differentially Private Logistic Regression} \label{subsec:logistic_results}

    First, we address a classic privacy-aware task: training a logistic regression model with DP-SGD on the widely-used Adult~\citep{adult_2} and Dutch~\cite{van2000integrating} benchmark datasets. 
    We consider HPO for batch size, learning rate, clipping threshold, and epochs and adopt the same search ranges as in~\cite{avent2020automatic}, which can be found in \Cref{tab:hyper-bounds}.
    The range of $\varepsilon$ is set as $[0.01, 0.5]$ and we fix $\delta=10^{-5}$.

     \begin{figure}[t]
        \centering
        \includegraphics[width=1\linewidth]{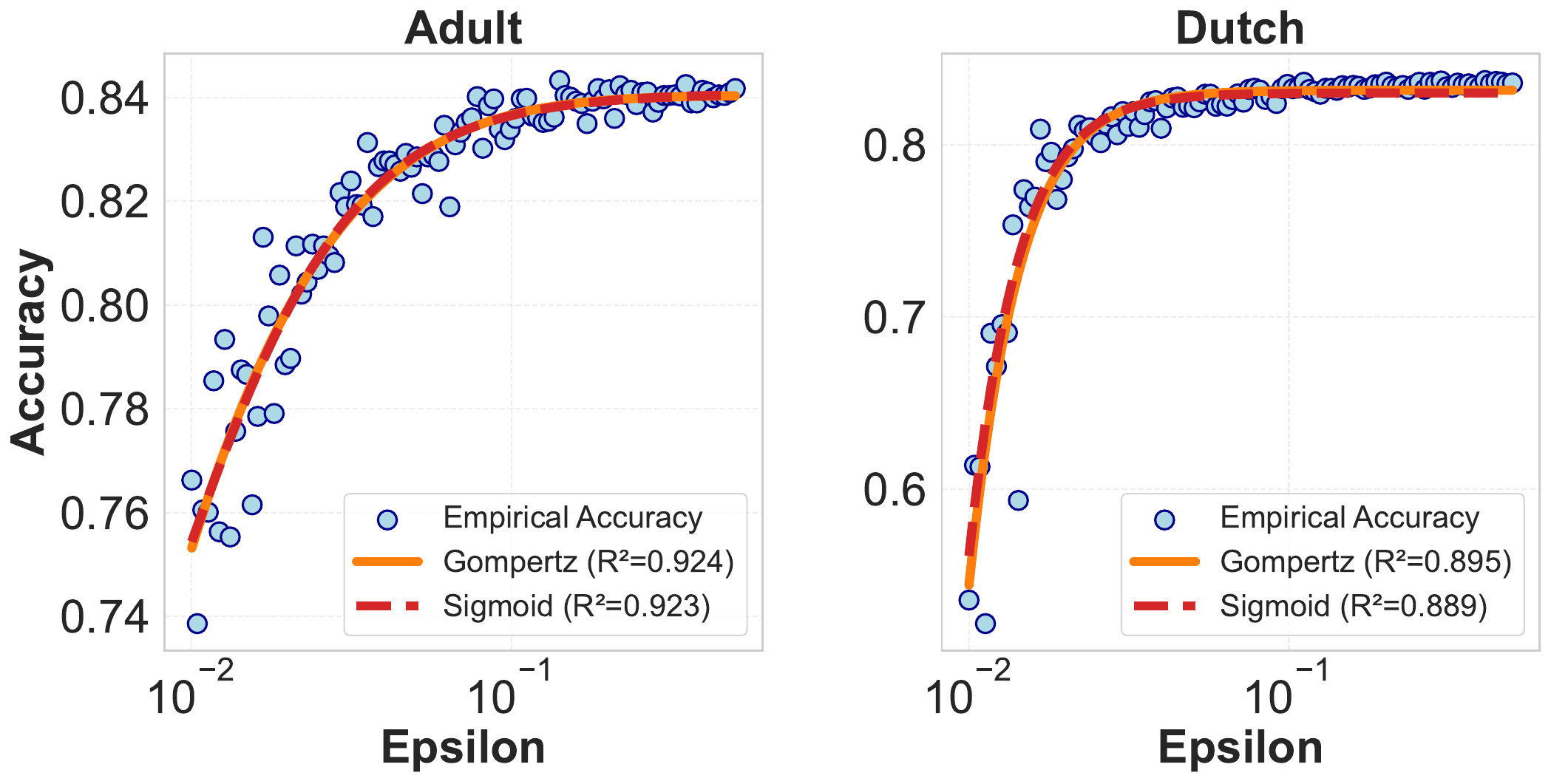}
        \caption{Sigmoid and Gompertz functions fitted to the empirical privacy-accuracy tradeoff for DP logistic regression on the Adult and Dutch datasets with $\delta=10^{-5}$. Both curves effectively capture the trend of the observed trade-off data.}
        \label{fig:logistic_fit}
    \end{figure}

    \begin{figure}[t]
        \centering
        \includegraphics[width=1\linewidth]{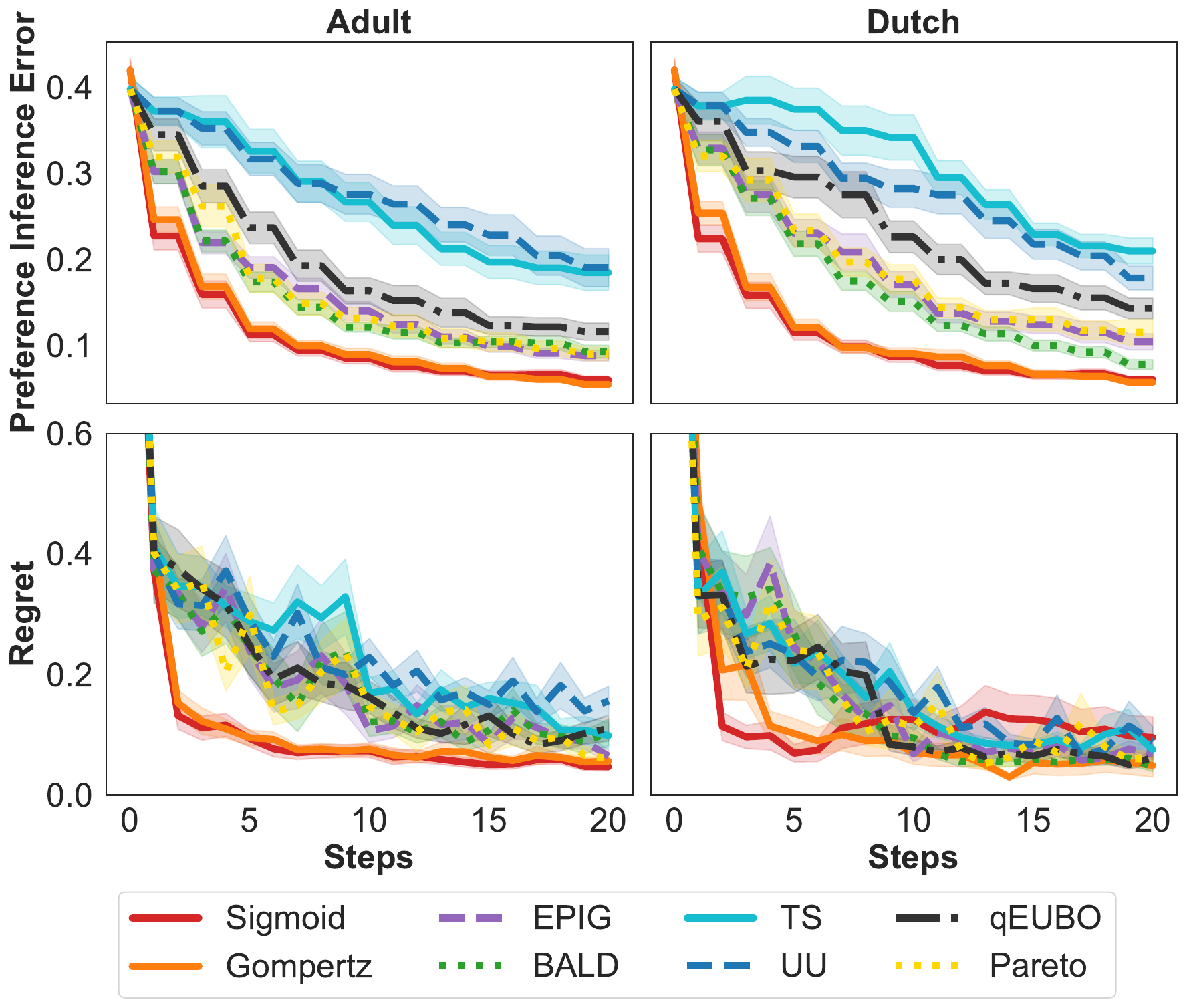}
        \caption{Results of DP logistic models. The top row shows the preference inference errors and the bottom row shows the regrets. Our methods (Sigmoid and Gompertz) have smaller errors and and regrets than baselines.}
         \label{fig:logistic_results}
    \end{figure}

    \Cref{fig:logistic_fit} presents the observed model accuracy (blue dots) after HPO as a function of the privacy budget, ${\varepsilon}$, for the Adult and Dutch datasets. 
    We fit two S-shaped surrogate models to this empirical data: the symmetric sigmoid function (red line) and the asymmetric Gompertz function (orange line). 
    Although these trade-off curves do not represent complete S-shapes -- they do not reach a lower asymptote even for very small epsilon values, the flexibility of the S-shape functions still ensure a high-quality fit.

    \Cref{fig:logistic_results} illustrates the performance of our proposed methods (using sigmoid and Gompertz surrogates) against the baselines, evaluated on two key metrics: preference inference error and regret.

    The top row of plots shows the preference inference error. 
    Across both datasets, our methods (red and orange lines) consistently achieve lower error rates than all baselines. 
    This demonstrates their better sample efficiency, as they can learn the decision-maker's preferences more accurately with fewer interactions.

    The bottom row shows the regret, which measures how effectively the methods find the decision-maker's optimal trade-off. 
    On the Adult dataset, our methods again show consistently lower regret. 
    On the Dutch dataset, while the sigmoid-based variant performs well in the initial steps, the Gompertz-based variant ultimately converges to a lower final regret, suggesting it forms a more accurate long-term model of the Pareto front.
        
    \begin{figure*}[t]
        \centering
        \includegraphics[width=1\linewidth]{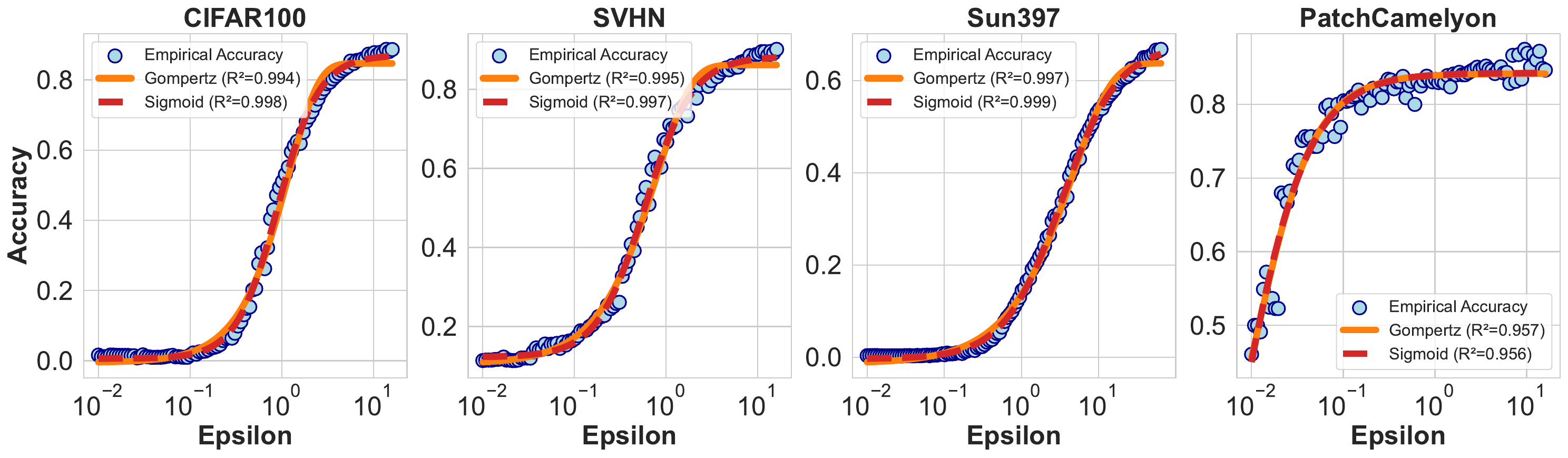}
        \caption{Sigmoid and Gompertz functions fitted to the empirical privacy-accuracy tradeoff for DP deep transfer learning models with $\delta=10^{-5}$. Both curves effectively capture the trend of the observed trade-off data.}
        \label{fig:finetuning_fit}
    \end{figure*}
    
    \begin{figure*}[t]
        \centering
        \includegraphics[width=1\linewidth]{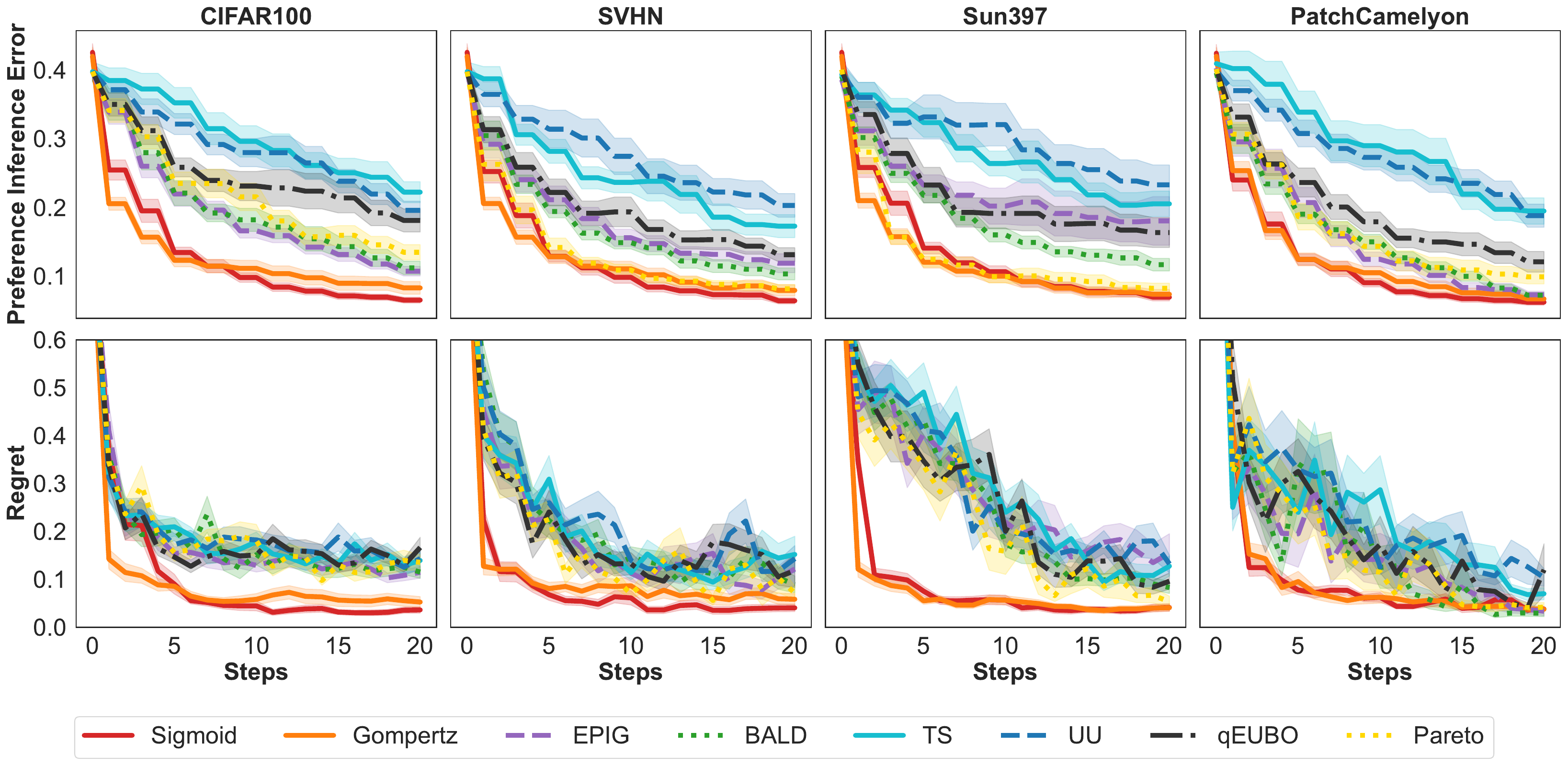}
        \caption{Results of DP deep transfer learning results. The top row shows the preference inference errors and the bottom row shows the regrets. Our methods (Sigmoid and Gompertz) have smaller errors and and regrets than baselines.}
        \label{fig:finetuning_results}
    \end{figure*}
\subsection{Finding the Optimal Trade-off in Differentially Private Deep Transfer Learning} \label{subsec:dp_results}

    In this section we conduct more advanced experiments using a Vision Transformer~\cite{dosovitskiy2021an} pretrained on ImageNet-21k \cite{ridnik2021imagenet} from the PyTorch Image Models library~\cite{rw2019timm} and fine-tuned on the target task.
    The backbone model contains 85.8 million parameters.
    We compare our method with baselines on four datasets: CIFAR-100~\cite{krizhevsky2009learning}, SVHN~\cite{netzer2011reading}, SUN~\cite{xiao2010sun} and PatchCamelyon~\cite{veeling2018rotation}.
    
    To enable faster training without compromising accuracy, we leverage DP-FiLM, as introduced by Tobaben et al. (\citeyear{tobaben_efficacy_2023}).
    In DP-FiLM, we freeze all model parameters except for the scale and bias of the normalization layers and the final classification layer.
    Following the DP-FiLM method, we initialize the classification layer weights to zero and make them trainable, resulting in 0.13--0.40\% of trainable parameters depending on the classification head size. The pretrained model expects $224 \times 224$ input images, so we resize images accordingly. 
    Training is performed in distributed mode across 4 AMD Radeon Instinct MI250X GPUs.
   
    We fix the number of epochs to 40--as it empirically gives close to state-of-the-art results for our datasets--and optimize the learning rate, the batch size, and the clipping threshold. 
    As we expect smaller values of the learning rate and the clipping threshold to give better results, we search for those in log space. 
    The search bounds for these hyperparameters are shown in \Cref{tab:hyper-bounds}.

    We use the Opacus~\cite{opacus} library to achieve DP guarantee.
    To facilitate hyperparameter tuning, we decouple the effect of the learning rate and the clipping threshold enabling the learning rate to absorb a fraction of the clipping threshold, as introduced by De et al. (\citeyear{de_unlocking_2022}).
    We always fix $\delta = 10^{-5}$.
    For accounting, we employ the PRV accountant~\cite{gopi_numerical_2021} that ships with Opacus.

    We also show the two S-shaped models fitting in \Cref{fig:finetuning_fit} for four datasets.
    The range of $\varepsilon$ is set as $[0.01, 16]$ for CIFAR100, SVHN and PatchCamelyon, and $[0.01, 64]$ for Sun397. 
    Both curves provide a good fit to the observed points and sigmoid fitting is slightly better than Gompertz fitting. 
    \Cref{fig:finetuning_results} shows that across all datasets, our methods have smaller errors in preference learning and regret.

    Our experiments demonstrate the robustness of our sigmoid-based approach across Pareto fronts with diverse shapes. 
    For instance, the trade-off curve for the PatchCamelyon dataset in \cref{fig:finetuning_fit} represents only the steep portion of an S-shaped curve, while the curve for Sun397 does not reach its upper asymptote even with a large privacy budget $\varepsilon=64$. 
    In both cases, our S-shaped model effectively captures the underlying trend, showcasing its flexibility.
    This highlights a key advantage of our framework: it does not require the full S-shaped curve to be present. 
    Decision-makers can define any specific range of epsilon that is relevant to their application, and our method can efficiently identify their preferred trade-off within that custom domain.

\subsection{Ablation Study} \label{subsec:ablation}
    
    In this section, we investigate the individual components that make up our method and show the performance of them.
    
\subsubsection{Pareto Front Estimation}

    This experiment investigates the sample efficiency of our proposed method--modeling the Pareto front directly--against the standard baseline approach of modeling the objective functions (recall \Cref{subsec:moo}).
    While our framework supports various S-shaped curves to model the front, we found that the Gompertz and sigmoid functions performed similarly across various datasets.
    Therefore, for clarity and illustration, we use the sigmoid function in the following experiments.
    We assume the true preferences weights are known and compare the regret (Equation \ref{eq:regret}) on finding the optimal trade-off of CIFAR100.
    We compare our method to other three baselines:
    1) our method, which selects values for constraint optimization with KG.
    2) \textbf{Sigmoid-random}: ablation approach which randomly selects constraints to optimize with.
    We add two (traditional) approaches that infer the Pareto front from estimating the objective functions:
    3) \textbf{GP-KG}: select candidates ($\bs{\theta}$) from the hyperparameter space that optimize KG.
    4) \textbf{GP-random}: randomly select candidates from the hyperparameter space.

    Results in the right pane of \Cref{fig:ablation} shows the comparisons.  
    Modeling the Pareto front requires clearly fewer samples than modeling the objectives. 
    The regret obtained when modeling the Pareto front with a sigmoid function (using either random sampling (green) or our acquisition function (red)) is consistently lower than when modeling the objectives directly.
    Note that while random sampling performs better in the initial steps, it converges to higher regret, whereas our method continues to reduce regret further.

\subsubsection{Preference Learning} \label{subsec:exp1}
    
    Here, we focus on the sample efficiency of preference learning given the true Pareto front, which is assumed to be known.
    In particular, we measure the preference inference error (Equation \ref{eq:pref-inf-err}) on $\bs{w}$, given feedback from the decision-maker queries. 
    In this task, we compare our method to the other three methods: 
    (1) Our method, which selects the next curve with KG.
    (2) \textbf{Random curve}: querying random hypothetical Pareto fronts, (3) \textbf{Pairs with KG}: pairwise comparisons picked with KG, and 
    (4) \textbf{Random pairs}: random pairwise comparisons.
    
    \begin{figure}[ht]
        \begin{center}
            \begin{minipage}[t]{0.23\textwidth}
                \centering
                \includegraphics[width=\textwidth]{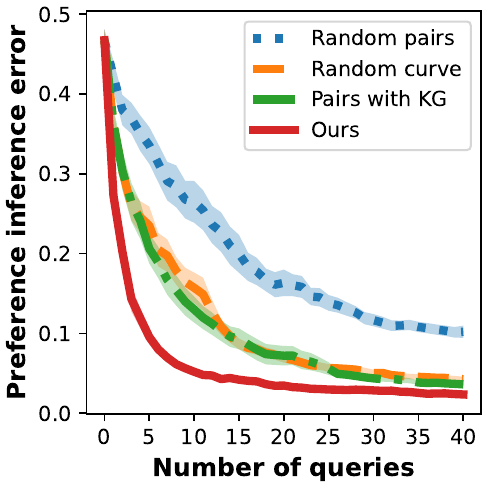}
            \end{minipage}
            \hfill
            \begin{minipage}[t]{0.23\textwidth}
                \centering
                \includegraphics[width=\textwidth]{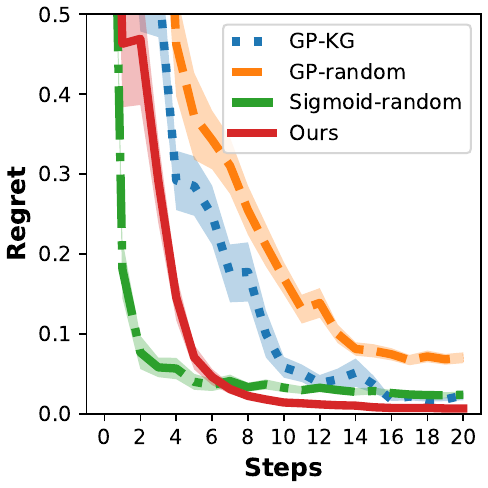}
            \end{minipage}
            \caption{Results of ablation studies. left: ablation study of different interaction mechanisms, where our method has the smallest inference errors. Right: ablation study of different methods to find the optimal trade-off, where ours converges to a smaller regret.}
            \label{fig:ablation}
        \end{center}
    \end{figure}

    Results in the left pane of \Cref{fig:ablation} confirm our expectations: random pairwise comparisons are the least informative, while pairs picked with KG perform similarly to random hypothetical Pareto fronts.
    Optimizing for informative fronts -- our method -- leads to the lowest inference error.

\subsection{Visualization of Utility Function} \label{subsec:visulization_u}

    \begin{figure}[t]
        \centering
        \includegraphics[width=1\linewidth]{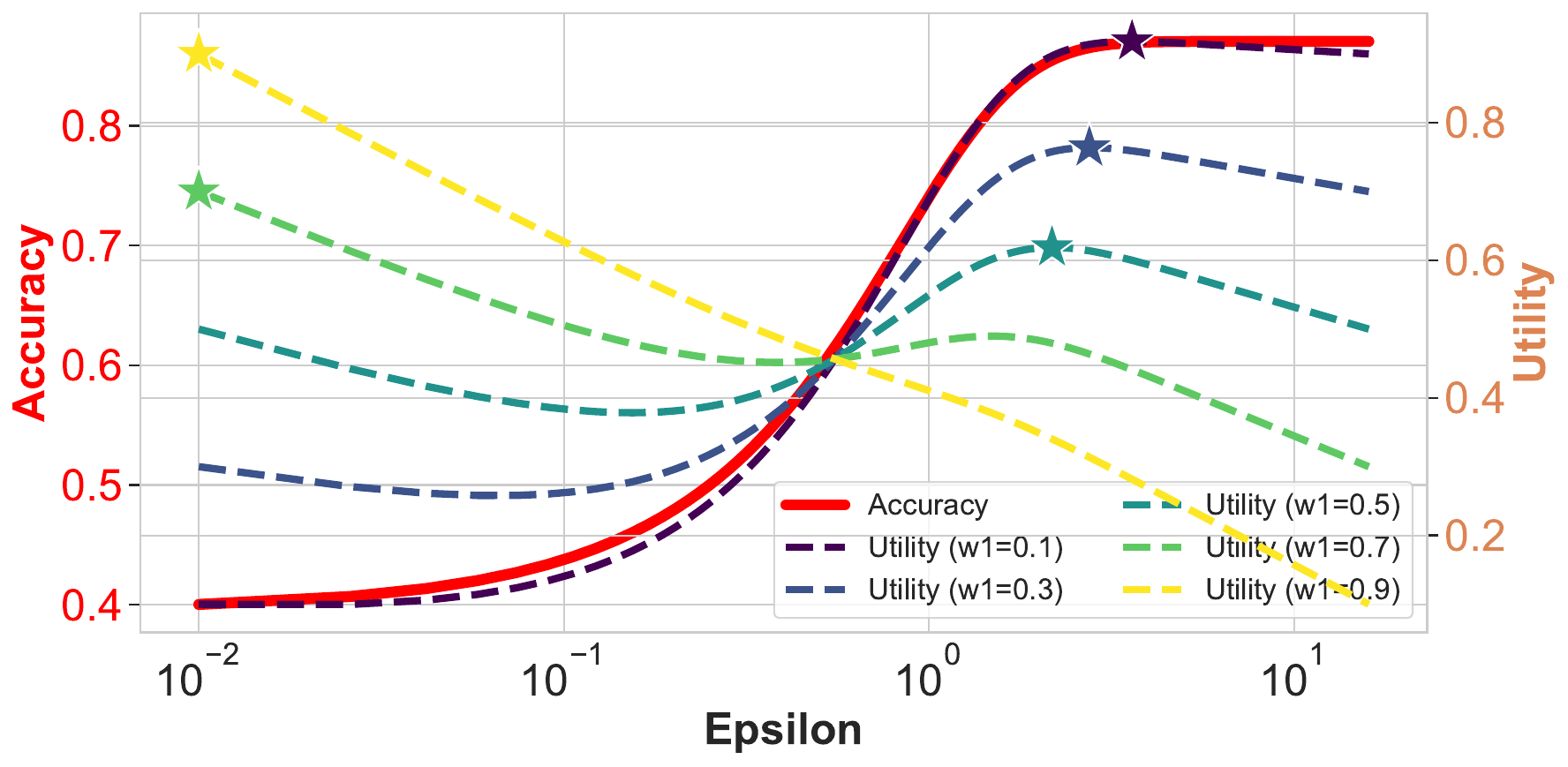}
        \caption{Visualization of linear utility function (Equation \ref{eq:linear_u}) for varying preference weights. The red solid curve presents the privacy-accuracy trade-off. Each star point marks the optimal privacy level that maximizes the utility for its corresponding weight, which fails to identify points in the crucial middle region of the curve. }
        \label{fig:linear_u}
    \end{figure}

    \begin{figure}[t]
        \centering
        \includegraphics[width=1\linewidth]{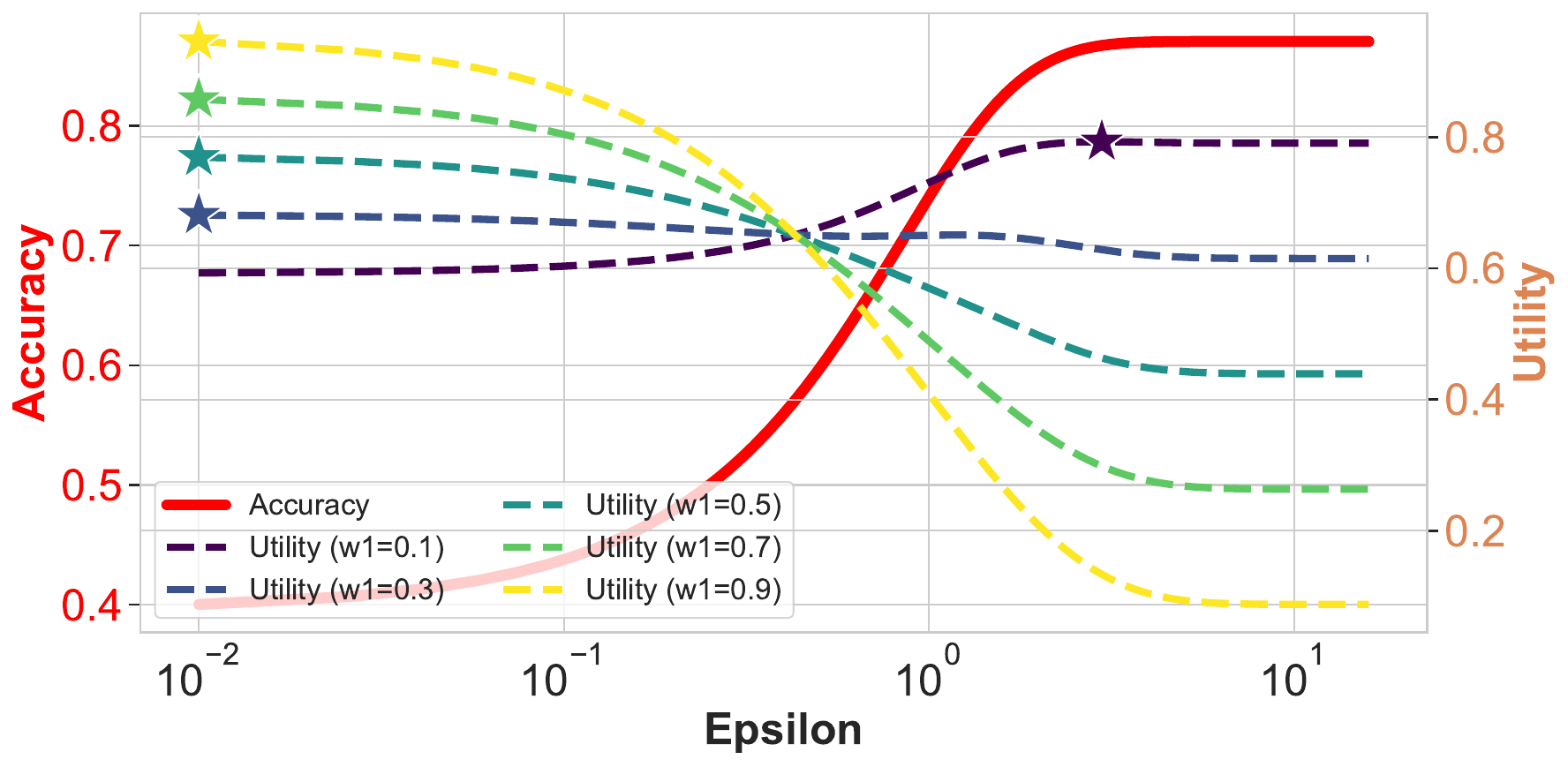}
        \caption{Visualization of linear utility function in~\cite{priyanshu2022efficient} for varying preference weights. The red solid curve presents the privacy-accuracy trade-off. Each star point marks the optimal privacy level that maximizes the utility for its corresponding weight, which fails to identify points in the crucial middle region of the curve.}
        \label{fig:linear_comparison}
     \end{figure}

    \begin{figure}[t]
        \centering
        \includegraphics[width=1\linewidth]{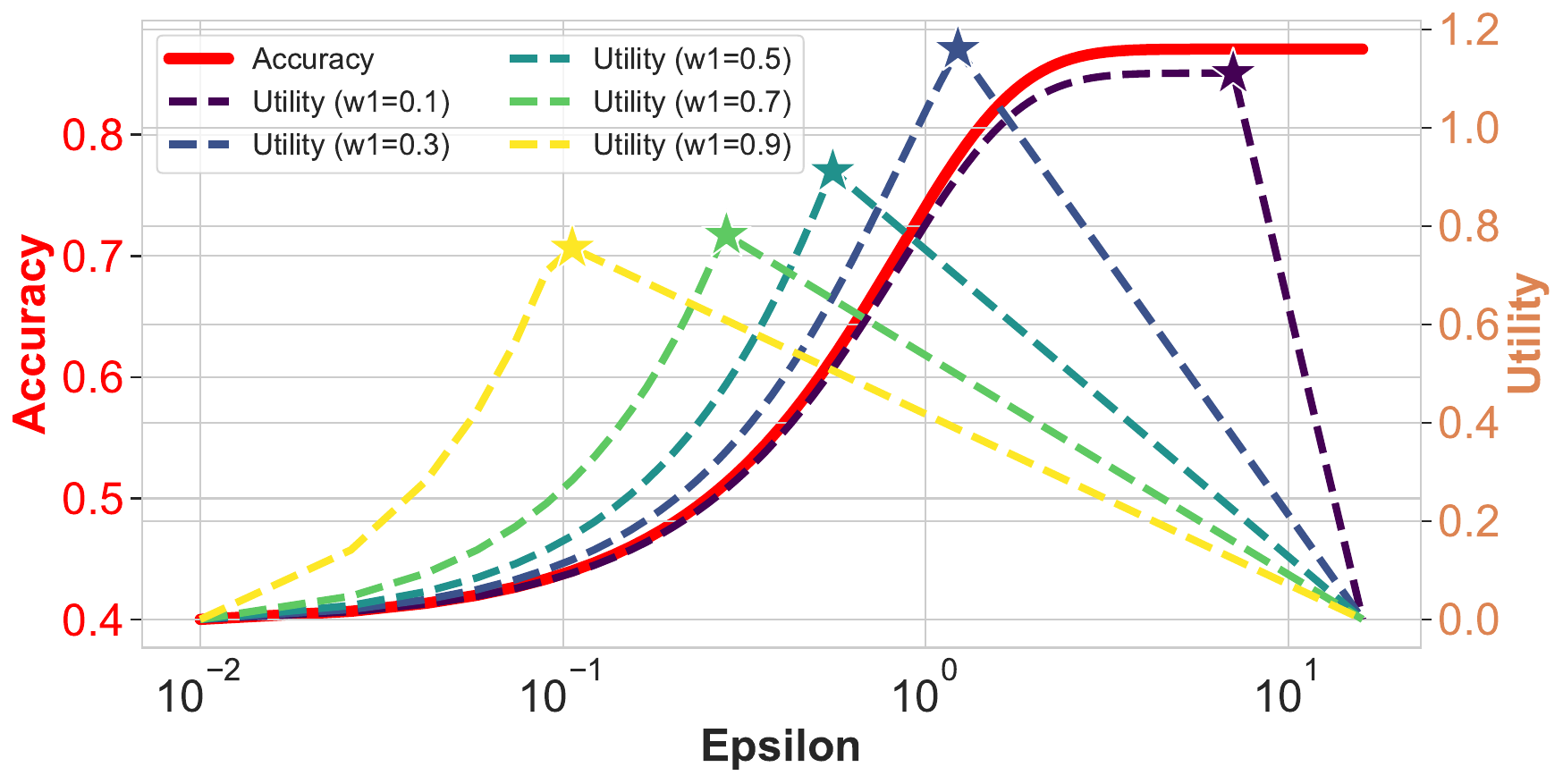}
        \caption{Visualization of Chebyshev utility function for varying preference weights. The red solid curve presents the privacy-accuracy trade-off. Each star point marks the optimal privacy level that maximizes the utility for its corresponding weight, which can successfully identify a diverse range of intermediate points.}
        \label{fig:cheby_u}
    \end{figure}

    In this section, we visualize the behavior of the Chebyshev utility function in comparison to linear utility functions when applied to the S-shaped privacy-accuracy Pareto front to suppoer our choice. 
    We have $w_1+w_2=1$ in all the utility functions and define $p=-\log{\varepsilon}$ as previously stated.

        \begin{align} 
        & U_{\text{Chebyshev}}(p, \alpha; \bs{w}) \\
        & = \min \left (\frac{1}{w_1}\frac{p-p_{\min}}{p_{\max}-p_{\min}}, \frac{1}{w_2}\frac{\alpha-\alpha_{\min}}{\alpha_{\max}-\alpha_{\min}}\right). \notag
        \end{align}
        
        \begin{align} 
        \label{eq:linear_u}
        &U_{\text{Linear}}(p, \alpha; \bs{w})\\
        &= w_1*\frac{p-p_{\min}}{p_{\max}-p_{\min}}+w_2*\frac{\alpha-\alpha_{\min}}{\alpha_{\max}-\alpha_{\min}}. \notag
        \end{align}

    In linear utility functions, it is possible that: 1) there exists a Pareto optimal solution that cannot be identified by any weight parameters, and 2) only one objective function with the highest weight is exclusively optimized~\cite{ozaki2024multi}. 
    \Cref{fig:linear_u} illustrates this failure of linear utility functions on the S-shaped Pareto front. 
    This problem is particularly acute for our S-shaped curve as the linear model fails to identify any of the nuanced, balanced trade-offs in the crucial middle region of the curve.
    \Cref{fig:linear_u} shows that for nearly all preference weights, the maximum utility is achieved at one of the two endpoints of the Pareto front. 

    We also visualize the linear utility proposed in \cite{priyanshu2022efficient} in \Cref{fig:linear_comparison}. 
    They consider the weighted linear combination of validation loss and $\varepsilon$ as the utility function for HPO. 
    As we explore the S-shaped trade-off curve between $\varepsilon$ and $\alpha$, we follow the idea in~\cite{avent2020automatic} and take classification error (=$1-\alpha$) as the metric for model performance. Then we have the following utility function according to~\cite{priyanshu2022efficient}:
    \begin{equation} 
        U_{\text{Linear}~\cite{priyanshu2022efficient}}({\varepsilon}, {\alpha}; \bs{w}) = w_1*e^{-{\varepsilon}}+w_2*e^{{\alpha}-1}.
    \end{equation}
    \Cref{fig:linear_comparison} shows the same issue with the \Cref{eq:linear_u} - for nearly all preference weights, the maximum utility is achieved at one of the two endpoints of the Pareto front.

    In stark contrast, \Cref{fig:cheby_u} demonstrates the suitability of the Chebyshev utility function for our framework. 
    We can see that as the decision-maker's preference weight changes, the peak of the utility curve shifts smoothly across the Pareto front. 
    The star points, which mark the optimal point for each weight, clearly show that the Chebyshev utility can identify a diverse range of intermediate points. 
    Therefore, a decision-maker can flexibly select any point on the S-shaped Pareto front, and our method can infer the corresponding preference weights that would make that choice optimal. 
    This creates a robust mapping from a decision-maker's action back to their latent preferences.

\section{Discussion} \label{sec:discussion}
    \paragraph{Setting the range of the privacy level}
    Our framework requires the decision-maker to pre-define a range of interest for the privacy level, $[p_{\min}, p_{\max}]$.
    There is no universally "correct" range; the appropriate bounds are entirely context-dependent and are dictated by the specific application's required trade-off.
    
    For instance, applications involving highly sensitive data, such as analyzing patient records in a healthcare setting, demand stringent privacy guarantees and would thus necessitate a range with a very low upper bound (e.g., $\varepsilon\leq 1$)~\cite{dankar2012application}. 
    Conversely, in scenarios where the individual user data is less sensitive and the primary goal is to improve a service for millions of users, a higher epsilon might be deemed acceptable. 
    A well-known example is Apple's use of differential privacy to learn from user typing behaviors to improve keyboard predictions, where a pragmatic trade-off allows for a higher $\varepsilon=16$ to maintain high accuracy~\cite{tang2017privacy}.
    
    Our method is designed to operate within such a user-defined range, regardless of its specific bounds. 
    The goal of our framework is not to prescribe a universal privacy standard, but rather to provide a principled tool that helps a decision-maker efficiently identify their optimal privacy-accuracy trade-off within their own pre-defined spectrum of acceptable options.

    \paragraph{Likelihood for observed accuracies} 
    In our current framework, we model the observed accuracy using a Gaussian likelihood with a constant variance $(\sigma^2)$. 
    This homoscedastic assumption implies that the reliability of our accuracy measurements is the same across the entire range of the privacy level, $p$.
    
    However, it is reasonable to suspect that this assumption could be refined. 
    One could hypothesize a heteroscedastic relationship, where the variance of the observed accuracy is itself a function of $p$. 
    For instance, at very large $p$ values, the large amount of injected noise makes the training process highly unstable, likely leading to a larger variance in the final accuracy across different runs. 
    Conversely, at small $p$ values, the training process is more stable and deterministic, suggesting a smaller variance.
    
    Future work could incorporate this by modeling the variance as a function of the privacy level. 
    Adopting such a heteroscedastic model could lead to a more accurate representation of the true data-generating process and the trade-off curve, then potentially improve the sample efficiency of the Pareto front exploration.

     \paragraph{Target Audience}
     The primary audience for our work is data practitioners and researchers with expertise in differential privacy. 
     Our framework is designed for decision-makers who understand the meaning of the privacy level and are tasked with the practical challenge of selecting an optimal privacy level to deploy the differentially private models.
     Our aim to provide a sample-efficient method to aid this complex decision-making process.

     While our primary focus is on expert users, the method has benefits for a broader audience. 
     By visualizing the entire trade-off curve, our approach can provide stakeholders and decision-makers without deep DP expertise an intuitive understanding of how accuracy degrades as privacy is strengthened. 
     However, we consider the important challenge of how to best communicate the nuances of differential privacy to a non-technical audience to be outside the scope of this paper. 
     This remains an active and valuable area of research, with notable work exploring how to translate formal privacy guarantees into understandable, practical terms for the general public~\cite{cummings2021need,nanayakkara2023chances}.

\section{Conclusion} \label{sec:conclusion}

    In this work, we proposed a more efficient approach for finding the optimal trade-off between privacy and accuracy with respect to a decision-maker's preference in differential privacy.
    To do so, we first derived both theoretical support and empirical evidence for the intuition that this trade-off is S-shaped.
    This finding, combined with the insight that hyperparameter tuning at a fixed privacy level naturally generates a solution on the Pareto front, allowed us to develop a Bayesian method for explicitly learning and modeling the Pareto-front.
    Additionally, utilizing this probabilistic model of the trade-off, we proposed a sample-efficient interaction scheme with a novel likelihood model to infer a decision-maker's preferences over the privacy-accuracy trade-off.
    Putting this all together, we showed our method vastly outperforms any state-of-the-art work in extensive empirical evaluations.

\section*{Acknowledgments}

This work was supported by the Finnish Ministry of Education and Culture and CSC - IT Centre for Science (Decision diary number OKM/10/524/2022), the Research Council of Finland (Flagship programme: Finnish Center for Artificial Intelligence, FCAI, Grant 356499 and Grant 359111), the Strategic Research Council at the Research Council of Finland (Grant 358247) as well as the European Union (Project 101070617) and UKRI Turing AI World-Leading Researcher Fellowship, EP/W002973/1. 
Views and opinions expressed are however those of the author(s) only and do not necessarily reflect those of the European Union or the European Commission. 
Neither the European Union nor the granting authority can be held responsible for them. 
This work has been performed using resources provided by the Aalto Science-IT Project from Computer Science IT, the CSC – IT Center for Science, Finland and the Finnish Computing Competence Infrastructure (FCCI).
We acknowledge CSC (Finland) for awarding this project access to the LUMI supercomputer, owned by the EuroHPC Joint Undertaking, hosted by CSC (Finland) and the LUMI consortium.

\bibliography{papers}

\begin{thebibliography}{66}
\providecommand{\natexlab}[1]{#1}
\providecommand{\url}[1]{\texttt{#1}}
\expandafter\ifx\csname urlstyle\endcsname\relax
  \providecommand{\doi}[1]{doi: #1}\else
  \providecommand{\doi}{doi: \begingroup \urlstyle{rm}\Url}\fi

\bibitem[Abadi et~al.(2016)Abadi, Chu, Goodfellow, McMahon, Mironov, Talwar, and Zhang]{abadi2016deep}
Abadi, M., Chu, A., Goodfellow, I., McMahon, H.~B., Mironov, I., Talwar, K., and Zhang, L.
\newblock Deep learning with differential privacy.
\newblock In \emph{Proceedings of the 2016 ACM SIGSAC Conference on Computer and Communications Security}, pp.\  308--318, New York, NY, USA, 2016. ACM, ACM.

\bibitem[Akiba et~al.(2019)Akiba, Sano, Yanase, Ohta, and Koyama]{akiba_optuna_2019}
Akiba, T., Sano, S., Yanase, T., Ohta, T., and Koyama, M.
\newblock Optuna: {A} {Next}-generation {Hyperparameter} {Optimization} {Framework}.
\newblock In \emph{Proceedings of the 25th {ACM} {SIGKDD} {International} {Conference} on {Knowledge} {Discovery} \& {Data} {Mining}}, {KDD} '19, pp.\  2623--2631, New York, NY, USA, July 2019. Association for Computing Machinery.
\newblock ISBN 978-1-4503-6201-6.
\newblock \doi{10.1145/3292500.3330701}.
\newblock URL \url{https://dl.acm.org/doi/10.1145/3292500.3330701}.

\bibitem[Astudillo \& Frazier(2020)Astudillo and Frazier]{astudillo2020multi}
Astudillo, R. and Frazier, P.
\newblock {Multi-attribute Bayesian optimization with interactive preference learning}.
\newblock In \emph{Proceedings of the 23rd International Conference on Artificial Intelligence and Statistics}, volume 108 of \emph{Proceedings of Machine Learning Research}, pp.\  4496--4507, Palermo, Sicily, Italy, 2020. PMLR.

\bibitem[Astudillo et~al.(2023)Astudillo, Lin, Bakshy, and Frazier]{astudillo2023qeubo}
Astudillo, R., Lin, Z.~J., Bakshy, E., and Frazier, P.
\newblock {qEUBO: A decision-theoretic acquisition function for preferential Bayesian optimization}.
\newblock In \emph{Proceedings of the 26th International Conference on Artificial Intelligence and Statistics}, volume 206 of \emph{Proceedings of Machine Learning Research}, pp.\  1093--1114, Valencia, Spain, 2023. PMLR.

\bibitem[Avent et~al.(2020)Avent, Gonz{\'a}lez, Diethe, Paleyes, and Balle]{avent2020automatic}
Avent, B., Gonz{\'a}lez, J., Diethe, T., Paleyes, A., and Balle, B.
\newblock {Automatic Discovery of Privacy--Utility Pareto Fronts}.
\newblock \emph{Proceedings on Privacy Enhancing Technologies}, 2020\penalty0 (4):\penalty0 441--461, 2020.

\bibitem[Becker \& Kohavi(1996)Becker and Kohavi]{adult_2}
Becker, B. and Kohavi, R.
\newblock {Adult}.
\newblock UCI Machine Learning Repository, 1996.
\newblock {DOI}: https://doi.org/10.24432/C5XW20.

\bibitem[Belakaria et~al.(2019)Belakaria, Deshwal, and Doppa]{belakaria2019max}
Belakaria, S., Deshwal, A., and Doppa, J.~R.
\newblock {Max-value entropy search for multi-objective Bayesian optimization}.
\newblock \emph{Advances in neural information processing systems}, 32, 2019.

\bibitem[Branke(2008)]{branke2008multiobjective}
Branke, J.
\newblock \emph{{Multiobjective optimization: Interactive and evolutionary approaches}}, volume 5252.
\newblock Springer Science \& Business Media, Berlin, Heidelberg, Germany, 2008.

\bibitem[Bun \& Steinke(2016)Bun and Steinke]{bun2016concentrated}
Bun, M. and Steinke, T.
\newblock Concentrated differential privacy: Simplifications, extensions, and lower bounds.
\newblock In \emph{Theory of cryptography conference}, pp.\  635--658, Berlin, Heidelberg, 2016. Springer, Springer Berlin Heidelberg.

\bibitem[Chaudhuri \& Monteleoni(2008)Chaudhuri and Monteleoni]{chaudhuri2008privacy}
Chaudhuri, K. and Monteleoni, C.
\newblock Privacy-preserving logistic regression.
\newblock \emph{Advances in neural information processing systems}, 21, 2008.

\bibitem[Chu \& Ghahramani(2005)Chu and Ghahramani]{chu2005preference}
Chu, W. and Ghahramani, Z.
\newblock {Preference learning with Gaussian processes}.
\newblock In \emph{Proceedings of the 22nd international conference on Machine learning}, pp.\  137--144, Vienna, Austria, 2005. PMLR.

\bibitem[Cummings et~al.(2021)Cummings, Kaptchuk, and Redmiles]{cummings2021need}
Cummings, R., Kaptchuk, G., and Redmiles, E.~M.
\newblock {" I need a better description": An Investigation Into User Expectations For Differential Privacy}.
\newblock In \emph{Proceedings of the 2021 ACM SIGSAC Conference on Computer and Communications Security}, pp.\  3037--3052, New York, NY, USA, 2021. ACM.

\bibitem[Dankar \& El~Emam(2012)Dankar and El~Emam]{dankar2012application}
Dankar, F.~K. and El~Emam, K.
\newblock The application of differential privacy to health data.
\newblock In \emph{Proceedings of the 2012 Joint EDBT/ICDT Workshops}, pp.\  158--166, New York, NY, USA, 2012. ACM.

\bibitem[Daulton et~al.(2023)Daulton, Balandat, and Bakshy]{daulton2023hypervolume}
Daulton, S., Balandat, M., and Bakshy, E.
\newblock Hypervolume knowledge gradient: a lookahead approach for multi-objective bayesian optimization with partial information.
\newblock In \emph{International Conference on Machine Learning}, pp.\  7167--7204. PMLR, 2023.

\bibitem[De et~al.(2022)De, Berrada, Hayes, Smith, and Balle]{de_unlocking_2022}
De, S., Berrada, L., Hayes, J., Smith, S.~L., and Balle, B.
\newblock Unlocking {High}-{Accuracy} {Differentially} {Private} {Image} {Classification} through {Scale}, June 2022.
\newblock URL \url{http://arxiv.org/abs/2204.13650}.
\newblock arXiv:2204.13650 [cs, stat].

\bibitem[De~Peuter et~al.(2024)De~Peuter, Zhu, Guo, Howes, and Kaski]{de2024preference}
De~Peuter, S., Zhu, S., Guo, Y., Howes, A., and Kaski, S.
\newblock Preference learning of latent decision utilities with a human-like model of preferential choice.
\newblock \emph{Advances in Neural Information Processing Systems}, 37:\penalty0 123608--123636, 2024.

\bibitem[Dong et~al.(2022)Dong, Roth, and Su]{dong2022gaussian}
Dong, J., Roth, A., and Su, W.~J.
\newblock Gaussian differential privacy.
\newblock \emph{Journal of the Royal Statistical Society Series B: Statistical Methodology}, 84\penalty0 (1):\penalty0 3--37, 2022.

\bibitem[Dosovitskiy et~al.(2021)Dosovitskiy, Beyer, Kolesnikov, Weissenborn, Zhai, Unterthiner, Dehghani, Minderer, Heigold, Gelly, Uszkoreit, and Houlsby]{dosovitskiy2021an}
Dosovitskiy, A., Beyer, L., Kolesnikov, A., Weissenborn, D., Zhai, X., Unterthiner, T., Dehghani, M., Minderer, M., Heigold, G., Gelly, S., Uszkoreit, J., and Houlsby, N.
\newblock An image is worth 16x16 words: Transformers for image recognition at scale.
\newblock In \emph{International Conference on Learning Representations}, 2021.
\newblock URL \url{https://openreview.net/forum?id=YicbFdNTTy}.

\bibitem[Dwork \& Rothblum(2016)Dwork and Rothblum]{dwork2016concentrated}
Dwork, C. and Rothblum, G.~N.
\newblock Concentrated differential privacy.
\newblock \emph{arXiv preprint arXiv:1603.01887}, abs/1603.01887, 2016.

\bibitem[Dwork et~al.(2006)Dwork, McSherry, Nissim, and Smith]{dwork2006calibrating}
Dwork, C., McSherry, F., Nissim, K., and Smith, A.
\newblock {Calibrating noise to sensitivity in private data analysis}.
\newblock In \emph{Theory of Cryptography: Third Theory of Cryptography Conference, TCC 2006, March 4-7, 2006. Proceedings 3}, pp.\  265--284, New York, NY, USA, 2006. Springer.

\bibitem[Dwork et~al.(2019)Dwork, Kohli, and Mulligan]{dwork2019differential}
Dwork, C., Kohli, N., and Mulligan, D.
\newblock {Differential privacy in practice: Expose your epsilons!}
\newblock \emph{Journal of Privacy and Confidentiality}, 9\penalty0 (2), 2019.

\bibitem[Elvira \& Martino(2022)Elvira and Martino]{elvira2022advances}
Elvira, V. and Martino, L.
\newblock {Advances in Importance Sampling}.
\newblock \emph{IEEE Signal Processing Magazine}, 39\penalty0 (3):\penalty0 39--60, 2022.

\bibitem[Frazier et~al.(2009)Frazier, Powell, and Dayanik]{frazier2009knowledge}
Frazier, P., Powell, W., and Dayanik, S.
\newblock {The knowledge-gradient policy for correlated normal beliefs}.
\newblock \emph{INFORMS journal on Computing}, 21\penalty0 (4):\penalty0 599--613, 2009.

\bibitem[F{\"u}rnkranz \& H{\"u}llermeier(2010)F{\"u}rnkranz and H{\"u}llermeier]{furnkranz2010preference}
F{\"u}rnkranz, J. and H{\"u}llermeier, E.
\newblock {Preference learning and ranking by pairwise comparison}.
\newblock In \emph{Preference learning}, pp.\  65--82. Springer, Berlin, Heidelberg, Germany, 2010.

\bibitem[Gelman et~al.(1995)Gelman, Carlin, Stern, and Rubin]{gelman1995bayesian}
Gelman, A., Carlin, J.~B., Stern, H.~S., and Rubin, D.~B.
\newblock \emph{Bayesian data analysis}.
\newblock Chapman and Hall/CRC, 1995.

\bibitem[Giovanelli et~al.(2024)Giovanelli, Tornede, Tornede, and Lindauer]{giovanelli2024interactive}
Giovanelli, J., Tornede, A., Tornede, T., and Lindauer, M.
\newblock {Interactive hyperparameter optimization in multi-objective problems via preference learning}.
\newblock In \emph{Proceedings of the AAAI Conference on Artificial Intelligence}, volume~38, pp.\  12172--12180, Palo Alto, CA, USA, 2024. AAAI Press.

\bibitem[Gonz{\'a}lez et~al.(2017)Gonz{\'a}lez, Dai, Damianou, and Lawrence]{gonzalez2017preferential}
Gonz{\'a}lez, J., Dai, Z., Damianou, A., and Lawrence, N.~D.
\newblock {Preferential Bayesian Optimization}.
\newblock In \emph{Proceedings of the 34th International Conference on Machine Learning}, volume~70 of \emph{Proceedings of Machine Learning Research}, pp.\  1282--1291, Sydney, Australia, 2017. PMLR, PMLR.

\bibitem[Gopi et~al.(2021)Gopi, Lee, and Wutschitz]{gopi_numerical_2021}
Gopi, S., Lee, Y.~T., and Wutschitz, L.
\newblock Numerical {Composition} of {Differential} {Privacy}.
\newblock In \emph{Advances in {Neural} {Information} {Processing} {Systems}}, volume~34, pp.\  11631--11642, Red Hook, New York, USA., 2021. Curran Associates, Inc.
\newblock URL \url{https://proceedings.neurips.cc/paper/2021/hash/6097d8f3714205740f30debe1166744e-Abstract.html}.

\bibitem[Gunantara(2018)]{gunantara2018review}
Gunantara, N.
\newblock {A review of multi-objective optimization: Methods and its applications}.
\newblock \emph{Cogent Engineering}, 5\penalty0 (1):\penalty0 1502242, 2018.

\bibitem[Jeon et~al.(2020)Jeon, Milli, and Dragan]{jeon2020reward}
Jeon, H.~J., Milli, S., and Dragan, A.
\newblock {Reward-rational (implicit) choice: A unifying formalism for reward learning}.
\newblock \emph{Advances in Neural Information Processing Systems}, 33:\penalty0 4415--4426, 2020.

\bibitem[Kinga et~al.(2015)Kinga, Adam, et~al.]{kinga2015method}
Kinga, D., Adam, J.~B., et~al.
\newblock {A method for stochastic optimization}.
\newblock In \emph{International conference on learning representations (ICLR)}, volume~5. California;, 2015.

\bibitem[Koskela \& Kulkarni(2023)Koskela and Kulkarni]{koskela2023practical}
Koskela, A. and Kulkarni, T.~D.
\newblock Practical differentially private hyperparameter tuning with subsampling.
\newblock \emph{Advances in Neural Information Processing Systems}, 36:\penalty0 28201--28225, 2023.

\bibitem[Krizhevsky et~al.(2009)Krizhevsky, Hinton, et~al.]{krizhevsky2009learning}
Krizhevsky, A., Hinton, G., et~al.
\newblock {Learning multiple layers of features from tiny images}.
\newblock Master's thesis, University of Toronto, 2009.

\bibitem[Lin et~al.(2022)Lin, Astudillo, Frazier, and Bakshy]{lin2022preference}
Lin, Z.~J., Astudillo, R., Frazier, P., and Bakshy, E.
\newblock {Preference exploration for efficient Bayesian optimization with multiple outcomes}.
\newblock In \emph{International Conference on Artificial Intelligence and Statistics}, pp.\  4235--4258. PMLR, 2022.

\bibitem[Liu \& Talwar(2019)Liu and Talwar]{liu2014private}
Liu, J. and Talwar, K.
\newblock Private selection from private candidates.
\newblock In Charikar, M. and Cohen, E. (eds.), \emph{Proceedings of the 51st Annual {ACM} {SIGACT} Symposium on Theory of Computing, {STOC} 2019, Phoenix, AZ, USA, June 23-26, 2019}, pp.\  298--309, New York, NY, USA, 2019. {ACM}.
\newblock \doi{10.1145/3313276.3316377}.
\newblock URL \url{https://doi.org/10.1145/3313276.3316377}.

\bibitem[Liu \& Bu(2025)Liu and Bu]{bu2025towards}
Liu, R. and Bu, Z.
\newblock {Towards hyperparameter-free optimization with differential privacy}.
\newblock In \emph{The Thirteenth International Conference on Learning Representations}, Singapore, 2025. ICLR.

\bibitem[Mavrotas(2009)]{mavrotas2009effective}
Mavrotas, G.
\newblock {Effective implementation of the $\varepsilon$-constraint method in multi-objective mathematical programming problems}.
\newblock \emph{Applied mathematics and computation}, 213\penalty0 (2):\penalty0 455--465, 2009.

\bibitem[Nanayakkara et~al.(2022)Nanayakkara, Bater, He, Hullman, and Rogers]{nanayakkara2022visualizing}
Nanayakkara, P., Bater, J., He, X., Hullman, J., and Rogers, J.
\newblock {Visualizing Privacy-Utility Trade-Offs in Differentially Private Data Releases}.
\newblock \emph{Proceedings on Privacy Enhancing Technologies}, 2022\penalty0 (3):\penalty0 521--541, 2022.

\bibitem[Nanayakkara et~al.(2023)Nanayakkara, Smart, Cummings, Kaptchuk, and Redmiles]{nanayakkara2023chances}
Nanayakkara, P., Smart, M.~A., Cummings, R., Kaptchuk, G., and Redmiles, E.~M.
\newblock {What are the chances? explaining the epsilon parameter in differential privacy}.
\newblock In \emph{32nd USENIX Security Symposium (USENIX Security 23)}, pp.\  1613--1630, Berkeley, CA, USA, 2023. USENIX Association.

\bibitem[Nanayakkara et~al.(2024)Nanayakkara, Kim, Wu, Sarvghad, Mahyar, Miklau, and Hullman]{nanayakkara2024measure}
Nanayakkara, P., Kim, H., Wu, Y., Sarvghad, A., Mahyar, N., Miklau, G., and Hullman, J.
\newblock {Measure-observe-remeasure: An interactive paradigm for differentially-private exploratory analysis}.
\newblock In \emph{2024 IEEE Symposium on Security and Privacy (SP)}, pp.\  1047--1064, Piscataway, NJ, USA, 2024. IEEE, IEEE.

\bibitem[Netzer et~al.(2011)Netzer, Wang, Coates, Bissacco, Wu, Ng, et~al.]{netzer2011reading}
Netzer, Y., Wang, T., Coates, A., Bissacco, A., Wu, B., Ng, A.~Y., et~al.
\newblock {Reading digits in natural images with unsupervised feature learning}.
\newblock In \emph{NIPS workshop on deep learning and unsupervised feature learning}, volume 2011, pp.\ ~4. Granada, 2011.

\bibitem[Ozaki et~al.(2024)Ozaki, Ishikawa, Kanzaki, Takeno, Takeuchi, and Karasuyama]{ozaki2024multi}
Ozaki, R., Ishikawa, K., Kanzaki, Y., Takeno, S., Takeuchi, I., and Karasuyama, M.
\newblock {Multi-Objective Bayesian Optimization with Active Preference Learning}.
\newblock In \emph{Proceedings of the AAAI Conference on Artificial Intelligence}, volume~38, pp.\  14490--14498, Palo Alto, CA, USA, 2024. AAAI Press.

\bibitem[Panda et~al.(2024)Panda, Tang, Mahloujifar, Sehwag, and Mittal]{pandanew}
Panda, A., Tang, X., Mahloujifar, S., Sehwag, V., and Mittal, P.
\newblock {A New Linear Scaling Rule for Private Adaptive Hyperparameter Optimization}.
\newblock In \emph{Proceedings of the 41st International Conference on Machine Learning}, volume 235 of \emph{Proceedings of Machine Learning Research}, pp.\  39074--39091, Vienna, Austria, 2024. PMLR.

\bibitem[Papernot \& Steinke(2022)Papernot and Steinke]{papernot2022hyperparameter}
Papernot, N. and Steinke, T.
\newblock Hyperparameter tuning with renyi differential privacy.
\newblock In \emph{The Tenth International Conference on Learning Representations, {ICLR} 2022, Virtual Event, April 25-29, 2022}. OpenReview.net, 2022.
\newblock URL \url{https://openreview.net/forum?id=-70L8lpp9DF}.

\bibitem[Pirouz \& Khorram(2016)Pirouz and Khorram]{pirouz2016computational}
Pirouz, B. and Khorram, E.
\newblock A computational approach based on the $\varepsilon$-constraint method in multi-objective optimization problems.
\newblock \emph{Adv. Appl. Stat}, 49\penalty0 (6):\penalty0 453--483, 2016.

\bibitem[Priyanshu et~al.(2022)Priyanshu, Naidu, Mireshghallah, and Malekzadeh]{priyanshu2022efficient}
Priyanshu, A., Naidu, R., Mireshghallah, F., and Malekzadeh, M.
\newblock {Efficient hyperparameter optimization for differentially private deep learning}.
\newblock In \emph{Proceedings of the 2022 ACM SIGSAC Conference on Computer and Communications Security}, pp.\  2455--2468, New York, NY, USA, 2022. ACM.

\bibitem[Rajkumar \& Agarwal(2012)Rajkumar and Agarwal]{rajkumar2012differential}
Rajkumar, A. and Agarwal, S.
\newblock A {{Differentially Private Stochastic Gradient Descent Algorithm}} for {{Multiparty Classification}}.
\newblock In \emph{{{AISTATS}}}, pp.\  933--941, La Palma, Canary Islands, January 2012.

\bibitem[Ridnik et~al.(2021)Ridnik, Baruch, Noy, and Zelnik]{ridnik2021imagenet}
Ridnik, T., Baruch, E.~B., Noy, A., and Zelnik, L.
\newblock {{ImageNet-21K Pretraining}} for the {{Masses}}.
\newblock In Vanschoren, J. and Yeung, S.-K. (eds.), \emph{The Thirty-fifth Conference on Neural Information Processing Systems, NeurIPS, Track on Datasets and Benchmarks}, 2021.

\bibitem[Robbins \& Monro(1951)Robbins and Monro]{robbins1951stochastic}
Robbins, H. and Monro, S.
\newblock A stochastic approximation method.
\newblock \emph{The annals of mathematical statistics}, 22:\penalty0 400--407, 1951.

\bibitem[Sander et~al.(2023)Sander, Stock, and Sablayrolles]{sander2023tan}
Sander, T., Stock, P., and Sablayrolles, A.
\newblock {{TAN Without}} a {{Burn}}: {{Scaling Laws}} of {{DP-SGD}}.
\newblock In \emph{The {{Fortieth International Conference}} on {{Machine Learning}}, {{ICML}} 2023}, pp.\  29937--29949. PMLR, 2023.

\bibitem[Smith et~al.(2023)Smith, Kirsch, Farquhar, Gal, Foster, and Rainforth]{smith2023prediction}
Smith, F.~B., Kirsch, A., Farquhar, S., Gal, Y., Foster, A., and Rainforth, T.
\newblock {Prediction-oriented bayesian active learning}.
\newblock In \emph{Proceedings of the 26th International Conference on Artificial Intelligence and Statistics}, volume 206 of \emph{Proceedings of Machine Learning Research}, pp.\  7331--7348, Valencia, Spain, 2023. PMLR, PMLR.

\bibitem[Song et~al.(2013)Song, Chaudhuri, and Sarwate]{song2013stochastic}
Song, S., Chaudhuri, K., and Sarwate, A.~D.
\newblock Stochastic gradient descent with differentially private updates.
\newblock In \emph{2013 {{IEEE Global Conference}} on {{Signal}} and {{Information Processing}}}, pp.\  245--248. IEEE, December 2013.
\newblock ISBN 978-1-4799-0248-4.
\newblock \doi{10.1109/GlobalSIP.2013.6736861}.

\bibitem[Suzuki et~al.(2020)Suzuki, Takeno, Tamura, Shitara, and Karasuyama]{suzuki2020multi}
Suzuki, S., Takeno, S., Tamura, T., Shitara, K., and Karasuyama, M.
\newblock {Multi-objective Bayesian optimization using Pareto-frontier entropy}.
\newblock In \emph{International conference on machine learning}, pp.\  9279--9288. PMLR, 2020.

\bibitem[Tang et~al.(2017)Tang, Korolova, Bai, Wang, and Wang]{tang2017privacy}
Tang, J., Korolova, A., Bai, X., Wang, X., and Wang, X.
\newblock {Privacy Loss in Apple's Implementation of Differential Privacy on MacOS 10.12}.
\newblock In \emph{Proceedings of the 2017 ACM SIGSAC Conference on Computer and Communications Security}, pp.\  183--199, New York, NY, USA, 2017. ACM.

\bibitem[Tobaben et~al.(2023)Tobaben, Shysheya, Bronskill, Paverd, Tople, Zanella-Beguelin, Turner, and Honkela]{tobaben_efficacy_2023}
Tobaben, M., Shysheya, A., Bronskill, J.~F., Paverd, A., Tople, S., Zanella-Beguelin, S., Turner, R.~E., and Honkela, A.
\newblock On the efficacy of {Differentially} {Private} {Few}-shot {Image} {Classification}.
\newblock \emph{Transactions on Machine Learning Research}, 2023:\penalty0 49, August 2023.
\newblock ISSN 2835-8856.
\newblock URL \url{https://openreview.net/forum?id=hFsr59Imzm}.

\bibitem[Ungredda \& Branke(2023)Ungredda and Branke]{ungredda2023elicit}
Ungredda, J. and Branke, J.
\newblock {When to elicit preferences in multi-objective Bayesian optimization}.
\newblock In \emph{Proceedings of the Companion Conference on Genetic and Evolutionary Computation}, pp.\  1997--2003, New York, NY, USA, 2023. ACM.

\bibitem[Ungredda et~al.(2022)Ungredda, Branke, Marchi, and Montrone]{ungredda2022single}
Ungredda, J., Branke, J., Marchi, M., and Montrone, T.
\newblock {Single interaction multi-objective Bayesian optimization}.
\newblock In \emph{Parallel Problem Solving from Nature -- PPSN XVII}, volume 13398 of \emph{Lecture Notes in Computer Science}, pp.\  132--145, Cham, Switzerland, 2022. Springer, Springer International Publishing.

\bibitem[Van~der Laan(2000)]{van2000integrating}
Van~der Laan, P.
\newblock Integrating administrative registers and household surveys.
\newblock \emph{Netherlands Official Statistics}, 15\penalty0 (2):\penalty0 7--15, 2000.

\bibitem[Veeling et~al.(2018)Veeling, Linmans, Winkens, Cohen, and Welling]{veeling2018rotation}
Veeling, B.~S., Linmans, J., Winkens, J., Cohen, T., and Welling, M.
\newblock {Rotation equivariant CNNs for digital pathology}.
\newblock In \emph{Medical Image Computing and Computer Assisted Intervention -- MICCAI 2018}, volume 11071 of \emph{Lecture Notes in Computer Science}, pp.\  210--218, Cham, Switzerland, 2018. Springer, Springer International Publishing.

\bibitem[Wightman(2019)]{rw2019timm}
Wightman, R.
\newblock {PyTorch Image Models}.
\newblock \url{https://github.com/rwightman/pytorch-image-models}, 2019.

\bibitem[Williams \& Rasmussen(1995)Williams and Rasmussen]{williams1995gaussian}
Williams, C. and Rasmussen, C.
\newblock Gaussian processes for regression.
\newblock \emph{Advances in neural information processing systems}, 8, 1995.

\bibitem[Winsor(1932)]{winsor1932gompertz}
Winsor, C.~P.
\newblock The gompertz curve as a growth curve.
\newblock \emph{Proceedings of the national academy of sciences}, 18\penalty0 (1):\penalty0 1--8, 1932.

\bibitem[Xiao et~al.(2010)Xiao, Hays, Ehinger, Oliva, and Torralba]{xiao2010sun}
Xiao, J., Hays, J., Ehinger, K.~A., Oliva, A., and Torralba, A.
\newblock {Sun database: Large-scale scene recognition from abbey to zoo}.
\newblock In \emph{2010 IEEE Computer Society Conference on Computer Vision and Pattern Recognition}, pp.\  3485--3492, Piscataway, NJ, USA, 2010. IEEE, IEEE.

\bibitem[Yamagata et~al.(2024)Yamagata, Oberkofler, Kaufmann, Bengs, H{\"u}llermeier, and Santos-Rodriguez]{yamagata2024relatively}
Yamagata, T., Oberkofler, T., Kaufmann, T., Bengs, V., H{\"u}llermeier, E., and Santos-Rodriguez, R.
\newblock {Relatively Rational: Learning Utilities and Rationalities Jointly from Pairwise Preferences}.
\newblock In \emph{ICML 2024 Workshop on Models of Human Feedback for AI Alignment}, volume 2024, Vienna, Austria, 2024. PMLR.

\bibitem[Yang et~al.(2019)Yang, Emmerich, Deutz, and B{\"a}ck]{yang2019multi}
Yang, K., Emmerich, M., Deutz, A., and B{\"a}ck, T.
\newblock Multi-objective bayesian global optimization using expected hypervolume improvement gradient.
\newblock \emph{Swarm and evolutionary computation}, 44:\penalty0 945--956, 2019.

\bibitem[Yousefpour et~al.(2021)Yousefpour, Shilov, Sablayrolles, Testuggine, Prasad, Malek, Nguyen, Ghosh, Bharadwaj, Zhao, Cormode, and Mironov]{opacus}
Yousefpour, A., Shilov, I., Sablayrolles, A., Testuggine, D., Prasad, K., Malek, M., Nguyen, J., Ghosh, S., Bharadwaj, A., Zhao, J., Cormode, G., and Mironov, I.
\newblock {Opacus: {U}ser-Friendly Differential Privacy Library in {PyTorch}}.
\newblock \emph{arXiv preprint arXiv:2109.12298}, abs/2109.12298, 2021.

\end{thebibliography}
\bibliographystyle{arXiv}


\appendix

\section{Sensitivity Analysis of $T$ in User Modeling} \label{appendix:user-model}
    Our proposed user modeling \Cref{eq:user_model}, which is shown below again,
    
        \begin{equation*}
           p(\bs{y^*} \mid \bs{\beta}, \bs{w}) = \frac{ \exp(U(\bs{y}^* ; \bs{w})/T) } { \sum_j^q \exp(U(\bs{y}_j;\bs{w})/T) },
        \end{equation*}
    
     where $T$ is the rationality coefficient. 
     While we take it as known in the preference learning procedure, inferring it along with other parameters is feasible~\cite{yamagata2024relatively,astudillo2023qeubo}. 
    Here we show the sensitivity analysis to show our model is not sensitive to the inferred $T$.
    
\begin{figure}[ht]
    \vskip 0.2in
    \begin{center}
    \centerline{\includegraphics[width=0.7\columnwidth]{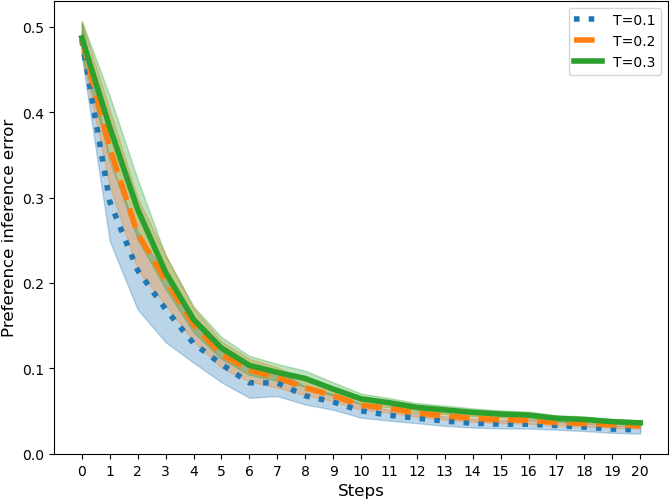}}
    \caption{Preference inference errors for different $T$ in the user model. There is no significant difference between different values.}
    \label{fig:sensitivity}
    \end{center}
    \vskip -0.2in
    \end{figure}

    We assume the Pareto front is known and simulate the decision-maker's action with $T_{\text{true}}=0.2$ and use $T=0.1, 0.2, 0.3$ in our user model to learn preferences. We see from \Cref{fig:sensitivity} that there is no significant difference to use different $T$ in our model. 
    
\section{Baselines Implementation} \label{appendix:baselines}

    When employing GP to model surrogates of privacy and accuracy in baselines, we use the same Matern kernel as in~\cite{avent2020automatic}: $k_M^{5/2}(\bs{x},\bs{x}')= (1+\frac{\sqrt{5}|\bs{x}-\bs{x}'|}{l}+\frac{5(\bs{x}-\bs{x}')^2}{3l^2})\exp(-\frac{\sqrt{5}|\bs{x}-\bs{x}'|}{l})$, where $l$ is the length-scale parameter to be optimized during inference.
    The number of initial points are 20, and every step 20 points are sampled based on EI acquisition function.

\section{Computation of Acquisition Functions} \label{appendix:compute_acq}
    For efficient approximation of the posteriors, we employ Monte-Carlo estimates with importance sampling~\cite{elvira2022advances}.
    We derive this for preference learning; the same approach is also applied to Pareto front estimation.
    
    Assume $\{\bs{y}^*_m,\bs{\beta}_m\}_{m=1}^M$ have been observed. When a new observation $(\bs{y}^*_{M+1}, \bs{\beta}_{M+1})$ comes, the target posterior is 
         \begin{align}\label{eq:IS}
           & p(\bs{w} \mid \{\bs{y}^*_m, \bs{\beta}_m\}_{m=1}^M, (\bs{y}^*_{M+1}, \bs{\beta}_{M+1}))\propto \notag\\
           &p(\bs{y}^*_{M+1} \mid \bs{\beta}_{M+1},\bs{w})\times\prod_{m=1}^M p(\bs{y}^*_{m} \mid \bs{\beta}_{m},\bs{w})\times p(\bs{w}) \propto \notag\\
           &p(\bs{y}^*_{M+1}\mid \bs{\beta}_{M+1},\bs{w})\times p(\bs{w} \mid \{\bs{y}^*_{m}, \bs{\beta}_{m}\}_{m=1}^M).
        \end{align}
    
    Importance sampling constitutes two steps: 
    \begin{enumerate}
        \item \textbf{Sampling:} $n_{\bs{w}}$ samples are simulated from $$\bs{w}_r\sim p(\bs{w} \mid \{\bs{y}^*_m,\bs{\beta}_m\}_{m=1}^M, \; r = 1,\dots, n_{\bs{w}}. $$
        \item \textbf{Weighting:} Each sample receives an associated importance weight given by
        \begin{align}\label{eq:weighting}
        p_r & = \frac{p(\bs{w}_r \mid (, \{\bs{y}^*_m,\bs{\beta}_m\}_{m=1}^M, (\bs{y}^*_{M+1}, \bs{\beta}_{M+1}))}{p(\bs{w}_r \mid \{\bs{y}^*_m, \bs{\beta}_m\}_{m=1}^M)} \notag \\
        & = p(\bs{y}^*_{M+1} \mid \bs{\beta}_{M+1},\bs{w}_r), 
        \end{align}
    \end{enumerate}
    which is exactly the likelihood of the new observations. 
    
    Similarly, in trade-off curve inference in DP case, the importance weight is
    $$
    p_s \approx \mathcal{N}(a_n  \mid  h(p_n; L, k, b, c), \sigma^2)\times d_n,
    $$
    where $d_n$ is a small interval. 
    
    The importance weights describe how representative the simulated samples are. The set of $n_{\bs{w}}$ weighted samples can be used to approximate the target posterior. We sample from the prior distribution of $\bs{w}$ and update weights as new observations arrive. Algorithm \ref{alg:IS_KG} details this importance sampling approach to KG computation.

     \begin{algorithm}
       \caption{Simulation and importance-sampling based computation of KG in preference learning.}
       \label{alg:IS_KG}
    \begin{algorithmic}
    \STATE {\bfseries Input:} 
    \FOR{All candidate curves}
        \FOR{$\text{sim}=1, \dots \text{Num}$ }
            \STATE Generate a simulated action based on the posterior of $\bs{w}$. 
            \STATE Update the posterior of $\bs{w}$ based on the simulation using (\ref{eq:weighting}).
            \STATE $\Delta^{\text{sim}} = U^*_{M+1,N} - U^*_{M,N}$.
       \ENDFOR
        \STATE Estimate KG by $\frac{1}{\text{Num}}\sum_{\text{sim}=1}^\text{Num} \Delta^{\text{sim}}.$
    \ENDFOR
       \STATE {\bfseries Output:} The curve with the highest KG value. 
    \end{algorithmic}
    \end{algorithm}

\end{document}